\documentclass[10pt,journal,compsoc]{IEEEtran}
\usepackage{color}
\usepackage{amsmath}
\usepackage{amsfonts}
\usepackage{mathrsfs}
\usepackage{amssymb}
\usepackage{url}
\usepackage{pgfplots}
\usepackage{subfigure}
\usepackage{multirow}
\usepackage{latexsym}
\usepackage{booktabs}
\usepackage{bm}
\usepackage{soul}
\usepackage{longtable}
\usepackage{arydshln}
\usepackage[encapsulated]{CJK}

\ifCLASSOPTIONcompsoc
\usepackage[nocompress]{cite}
\else
\usepackage{cite}
\fi
\ifCLASSINFOpdf
\else
\fi
\begin{document}
	
	\title{\textcolor{black}{Text Compression-aided Transformer Encoding}} 
	
	\author{Zuchao Li, Zhuosheng Zhang, Hai Zhao$^*$, Rui Wang$^*$, Kehai Chen, Masao Utiyama, and Eiichiro Sumita
		% <-this % stops a space
		\IEEEcompsocitemizethanks{\IEEEcompsocthanksitem{This paper was partially supported by National Key Research and Development Program of China (No. 2017YFB0304100), Key Projects of National Natural Science Foundation of China (U1836222 and 61733011), Huawei-SJTU long term AI project, Cutting-edge Machine Reading Comprehension and Language Model (Corresponding author: Hai Zhao and Rui Wang).}
		\IEEEcompsocthanksitem{Z. Li, Z. Zhang, H. Zhao, and R. Wang are with the Department of Computer Science and Engineering, Shanghai Jiao Tong University, and also with Key Laboratory of Shanghai Education Commission for Intelligent Interaction and Cognitive Engineering, Shanghai Jiao Tong University, and also with MoE Key Lab of Artificial Intelligence, AI Institute, Shanghai Jiao Tong University. E-mail: \{charlee, zhangzs, \}@sjtu.edu.cn, zhaohai@cs.sjtu.edu.cn, wangrui.nlp@gmail.com. \protect\\ K. Chen, M. Utiyama, and E. Sumita are with National Institute of Information and Communications Technology (NICT), Kyoto, Japan, 619-0289. E-mail: \{khchen, mutiyama, eiichiro.sumita\}@nict.go.jp. Tel:+81-0774-6986.}
    	\IEEEcompsocthanksitem{Part of this work was finished when Z. Li and Z. Zhang visited National Institute of Information and Communications Technology (NICT) and R. Wang was with NICT.}}
	}
	
	% The paper headers
	\markboth{IEEE Transactions on Pattern Analysis and Machine Intelligence, March~2020}%
	{Li \MakeLowercase{\textit{et al.}}: Text Compression-aided Transformer Encoding}

	\IEEEtitleabstractindextext{%
		\begin{abstract}
			Text encoding is one of the most important steps in Natural Language Processing (NLP). 
			It has been done well by the self-attention mechanism in the current state-of-the-art Transformer encoder, which has brought about significant improvements in the performance of many NLP tasks.
			Though the Transformer encoder may effectively capture general information in its resulting representations, the backbone information, meaning the gist of the input text, is not specifically focused on.
			In this paper, we propose explicit and implicit text compression approaches to enhance the Transformer encoding and evaluate models using this approach on several typical downstream tasks that rely on the encoding heavily.
			Our explicit text compression approaches use dedicated models to compress text, while our implicit text compression approach simply adds an additional module to the main model to handle text compression.
			We propose three ways of integration, namely backbone source-side fusion, target-side fusion, and both-side fusion, to integrate the backbone information into Transformer-based models for various downstream tasks.
			Our evaluation on benchmark datasets shows that the proposed explicit and implicit text compression approaches improve results in comparison to strong baselines.
			We therefore conclude, when comparing the encodings to the baseline models, text compression helps the encoders to learn better language representations.
		\end{abstract}
		
		% Note that keywords are not normally used for peerreview papers.
		}

	% make the title area
	\maketitle

	\IEEEdisplaynontitleabstractindextext

	\IEEEpeerreviewmaketitle

	\IEEEraisesectionheading{\section{Introduction}\label{sec:introduction}}
	
	\IEEEPARstart{T}ext encoding plays an important role in NLP, especially in Natural Language Understanding (NLU). 
	NLU, which requires computers to read and understand human natural language texts, has been a long-standing goal of machine intelligence since the pioneering work of distributional text encoding by a neural network starting in 2003 \cite{bengio2003neural,mikolov2013distributed}, etc.
	Recently, statistical contextualized text encoding that effectively integrates contextual features and language model training objectives have set a series of state-of-the-art benchmarks in several NLP tasks \cite{devlin-etal-2019-bert, peters-etal-2018-deep,radford2018improving}. 
	So far, most of these existing works consider text encoding only from a language symbol distributional standpoint.
	Namely, these works use either token (word/subword/character) embedding or sentence-level encoders rooted in statistical co-occurrence. But some explicit experience of human understanding of the language is less taken into consideration to enrich the resulting representation. 
	
	In human languages, a sentence is a minimal unit that may delivers a complete piece of knowledge \cite{halliday2014introduction,ukessays2018distinguishing}.
	Intuitively, when a human reads a sentence, especially a long sentence, he/she often selectively focuses on some important words with basic sentence meaning and re-reads the sentence to understand its meaning completely; that is, some words (i.e., backbone words) are more important for understanding the basic meaning of this sentence than others (i.e., detail words). 
	In the state-of-the-art Transformer-based encoders, self-attention mechanisms effectively capture the general information in the input sentence/passage \cite{dou-etal-2018-exploiting,wang-etal-2019-exploiting,yang2019context}; \textcolor{black}{however, it is difficult to distinguish which information in the input is really salient for encoding.}
	Taking examples from the inputs of WMT14 English-to-German translation task (sentence-level) and SQuAD 2.0 reading comprehension task (paragraph-level) in Table \ref{tab:sc_example},
	we manually annotate the text's basic meaning as a sequence of words shorter than that of the original input and call this ``backbone" information.
	Obviously, these words in the backbone annotation contain more important information for human understanding than do the other words in the text.
	
	\begin{table*}[t]
		\centering 
		\caption{\textcolor{black}{Examples of text and backbone information.}}\label{tab:sc_example}
		\begin{tabular}{lp{14cm}}
			\toprule
			{\bf Sentence }
			& \textit{Both the {\bf \textit{US}} {\bf \textit{authorities}} {\bf \textit{and}} the {\bf \textit{Mexican}} security {\bf \textit{forces}} are engaged in an ongoing {\bf \textit{battle against}} the {\bf \textit{drug cartels}}.} \\ 
			\midrule
			{\bf Sentence-level Backbone} & \textit{US authorities and Mexican forces battle against drug cartels} \\
			\midrule
			{\bf Passage }
			&  \textit{{\bf \textit{The Normans}} (Norman: Nourmands; French: Normands; Latin: Normanni)  {\bf \textit{were the people who in the 10th and 11th centuries gave their name to Normandy, a region in France.}}  {\bf \textit{They were descended from Norse}} ("Norman" comes from "Norseman")  {\bf \textit{raiders and pirates from Denmark, Iceland and Norway who, under their leader Rollo}}, agreed to swear fealty to King Charles III of West Francia. Through generations of assimilation and mixing with the native Frankish and Roman-Gaulish populations, their descendants would gradually merge with the Carolingian-based cultures of West Francia.  {\bf \textit{The distinct cultural and ethnic identity of the Normans emerged initially in the first half of the 10th century}, and it continued to evolve over the succeeding centuries.}} \\ 
			\midrule
			{\bf Paragraph-level Backbone} & \textit{The Normans were the people who in the 10th and 11th centuries gave their name to Normandy, a region in France. They were descended from Norse raiders and pirates from Denmark, Iceland and Norway who, under their leader Rollo. The distinct cultural and ethnic identity of the Normans emerged initially in the first half of the 10th century.} \\
			\bottomrule
		\end{tabular}
	\end{table*}
	
	\textcolor{black}{We argue that such backbone information is helpful for text encoding since it is underutilized by Transformer encoders. 
	In other words, there are often redundancies in self-attentional token-level representations generated by Transformer encoders because semantically dominant words treated as being equal to less semantically significant tokens. 
	In this paper, this process of obtaining text backbone information is called text compression.}
	\textcolor{black}{We propose utilizing an Explicit Text Compression (ETC) approach (text summarization) and a novel Implicit Text Compression (ITC) approach to enhance Transformer-based models' text encoding with backbone information.}
	To this end, we first build three text compression settings, including supervised, unsupervised, and semi-supervised, to accommodate the needs of various scenarios where the compression sequences or features are learnt from the input.
	\textcolor{black}{Then, three methods of backbone fusion are proposed (\textbf{BEF}, \textbf{BDF}, and \textbf{BBF}), to integrate the backbone features into the Transformer encoder to improve the text encoding.}
	
	Empirical results on the widely used WMT14 English-to-German and English-to-French translation tasks and SQuAD 2.0 and RACE reading comprehension tasks show that the proposed approaches improve the performances of several NLP tasks over the strong and even state-of-the-art baselines.
	
	In addition, in order to further analyze the improvement source of proposed text compression approaches, we trained the baseline and the ETC-aided model on the Natural Language Inference (NLI) dataset and evaluated them using a controlled evaluation set, HANS (Heuristic Analysis for NLI Systems) \cite{mccoy-etal-2019-right}, which can effectively assess how much the model depends on the heuristic rules of the dataset and how much actual improvement it produces. The evaluation results show that our proposed ETC-aided model relies less on the heuristic rules. 
	\textcolor{black}{Using 10 linguistic tasks, further probing evaluation on the sentence embeddings derived from the NLI model verifies that the representations with the aids of text compression are improved.}

	\section{Backgrounds}\label{sec:backgrounds}
	
	\subsection{\textcolor{black}{Text Representations Learning}}%
	
	Recently, deep contextual Language Models (LMs) have been shown effective for text representation learning (i.e., text encoding) and achieving state-of-the-art results in a series of flagship NLU tasks. 
	Some prominent examples are Embedding from Language models (ELMo) \cite{peters-etal-2018-deep}, Generative Pre-trained Transformer (OpenAI GPT)  \cite{radford2018improving}, Bidirectional Encoder Representations from Transformers (BERT) \cite{devlin-etal-2019-bert}, Generalized Autoregressive Pre-training (XLNet) \cite{yang2019xlnet}, and A Lite BERT for Self-supervised Learning of Language Representations (ALBERT) \cite{lan2019albert}. 
	Providing fine-grained contextual text embedding, these pre-trained LMs can be easily applied to downstream models by serving as the encoder or being used for finetuning. 
	
	A number of studies have found that deep learning models might not really understand natural language \cite{mudrakarta-etal-2018-model} and be vulnerable to adversarial attacks \cite{jia-liang-2017-adversarial}. Deep learning models pay great attention to non-significant words and ignore important ones, which suggests that the current NLU models suffer from insufficient contextual semantic representation and learning. Natural language understanding tasks require a comprehensive understanding of natural languages and the ability to do further inference and reasoning. A common trend among NLU studies is that models are becoming more and more sophisticated with stacked attention mechanisms or are training from a larger amount of data, resulting in an explosive growth of computational cost. 
	
	Distributed embeddings have been widely used as a standard part of NLP models due to their ability to capture the local co-occurrence of words from large-scale unlabeled text \cite{mikolov2013distributed}; however, these approaches for learning word vectors only involve a single, context-independent embedding for each word with little consideration of contextual encoding at the sentence level. Thus, recently introduced contextual language models including ELMo, GPT, BERT, XLNet, and ALBERT fill this contextual gap by strengthening the contextual sentence modeling and consequently produce better representations. Among these models, BERT pioneered a powerful pre-training objective, \emph{masked language modeling}, which allows capturing both sides of context - the text preceding and following a word. Besides, BERT also introduces a \emph{next sentence prediction} objective that jointly pre-trains text-pair representations. 

	\subsection{Typical NLU Tasks}
	The human ability to understand language is general, flexible, and robust. Typical NLU tasks require models that can learn to represent linguistic knowledge in a way that facilitates sample-efficient learning and effective knowledge-transfer across tasks. Presently, the neural network-based NLP framework consistently achieves new levels of quality and has become the dominating approach for NLP tasks such as machine translation, machine reading comprehension, sentiment analysis, and textual entailment.
	
	In this work, we choose three tasks as research objectives: Neural Machine Translation (NMT), Machine Reading Comprehension (MRC), and Natural Language Inference (NLI). It is worth noting that in the era of traditional rule-based and phrase-based statistical machine translation, translation relied on rules and statistical probability and did not possibly constitute true language understanding; therefore, we do not consider statistical machine translation to fall under the category of NLU. On the contrary, NMT, which has shown much recent success, does not rely on such rules and thus requires a much more comprehensive understanding of the language used. Therefore, we argue that in addition to being a task of natural language generation, NMT can also  be regarded as a task of NLU, as it reflects the quality of understanding in two or more languages with the intuition that the better the understanding, the higher the quality of translation.
	
	NMT is popularly implemented as an encoder-decoder framework \cite{vaswani2017attention} in which the encoder handles source sentence representation.
	Typically, the input sentence is encoded as a contextualized source representation using deep learning networks.
	By feeding this to the decoder, the source representation is used to learn time-step dependent context vectors for predicting target translations~\cite{bojar-etal-2018-findings}. 
	
	MRC is a field of question-answering in which computers understand passages and answer related questions. Most of the previous MRC models are composed of two modules: an encoder and a pointing module. A passage (paragraph) is encoded as a contextualized representation, and the starting and ending positions of answer phrases are determined in the pointing module.
	
	The data for Multi-Genre Natural Language Inference (MNLI) \cite{williams-etal-2018-broad} consists of a crowd-sourced collection of sentence pairs with textual entailment annotations. Given a premise sentence and a hypothesis sentence, the task is to predict whether the premise entails the hypothesis (entailment), contradicts the hypothesis (contradiction), or neither (neutral).

	\subsection{Transformer Encoder}\label{encoder}
	
	Recurrent Neural Networks (RNNs) (or Long Short-Term Memory, LSTMs) have been established as advanced approaches to language modeling. Before 2017, they were the most optimal architecture for a sequence encoder. Because of their sequential nature, recurrent encoders pose a problem for learning long-term memory dependencies due to gradient explosion and a lack of parallelization during training. Vaswani et al. \cite{vaswani2017attention} introduced a novel architecture called Transformer, which provides a more structured memory for handling long-term dependencies in text, resulting in robust performance across diverse tasks.
	
	A Transformer encoder fully relies on self-attention networks (\textbf{SANs}) to encode an input sequence into a latent representation space. Formally, a input language sequence $x$=$\{x_1, \cdots, x_J\}$ of length $J$ is first mapped into embedding vectors. Then, the vectors and its position embeddings add up to form the input representation $v_x=\{v^x_1, \cdots, v^x_J\}$. The input representation $v_x$ is then packed into a query matrix $\textbf{Q}_x$, a key matrix $\textbf{K}_x$, and a value matrix $\textbf{V}_x$.
	For the SAN-based encoder, the self-attention sub-layer is first performed over $\textbf{Q}$, $\textbf{K}$, and $\textbf{V}$ to the matrix of outputs as:
	\begin{equation}
	\textup{SelfAtt}(\textbf{Q}, \textbf{K}, \textbf{V})=\textup{Softmax}(\frac{\textbf{Q}\textbf{K}^{T}}{\sqrt{d_{model}}})\textbf{V},\\
	\label{eq1:self-attention}
	\end{equation}
	where $d_{model}$ represents the dimensions of the model. 
	Generally, the self-attention function is further refined as multi-head self-attention to jointly consider information from different representation subspaces at different positions: 
	\begin{equation}
	\begin{split}
	&  \textup{MultiHead}(\textbf{Q}, \textbf{K}, \textbf{V}) =\textup{Concat}(\textup{head}_{1}, \cdots, \textup{head}_H)\textbf{\textit{W}}^{O},\\
	& \;\;\;\;\;\;\;\;\;\;\;\;\;\;\;\;\;\;\;\;\;\textup{head}_h=\textup{SelfAtt}(\textbf{Q}\textbf{W}_{h}^{Q},\textbf{k}\textbf{W}_{h}^{K},\textbf{V}\textbf{W}_{h}^{V}), 
	\end{split}
	\label{eq2:multiHead_self-attention}
	\end{equation}
	where the projections are parameter matrices $\textbf{W}_{h}^{Q}$$\in$$\mathbb{R}^{d_{model}\times d_k}$,  $\textbf{W}_{h}^{K}$$\in$$\mathbb{R}^{d_{model}\times d_k}$, $\textbf{W}_{h}^{V}$$\in$$ \mathbb{R}^{d_{model}\times d_v}$, and $\textbf{W}^{O}$$\in$$\mathbb{R}^{d_{v}\times d_{model}}$.
	For example, a common setup is $H$=8 heads, $d_{model}$ is 512, and $d_k$=$d_v$=512/8=64.
	A position-wise feed-forward network (FFN) layer is applied over the output of multi-head self-attention, and is then added with the matrix $\textbf{V}$ to generate the final source representation $H_{x}$=$\{H^{x}_1, \cdots, H^{x}_J\}$:
	\begin{equation}
	H_{x} = \textup{FFN}(\textup{MultiHead}(\textbf{Q}, \textbf{K}, \textbf{V}))+\textbf{V}.
	\label{eq3:PositionWiseFeedForward}
	\end{equation}
	
	\section{Overview}
	
\begin{figure}
		\centering
		\includegraphics[width=0.5\textwidth]{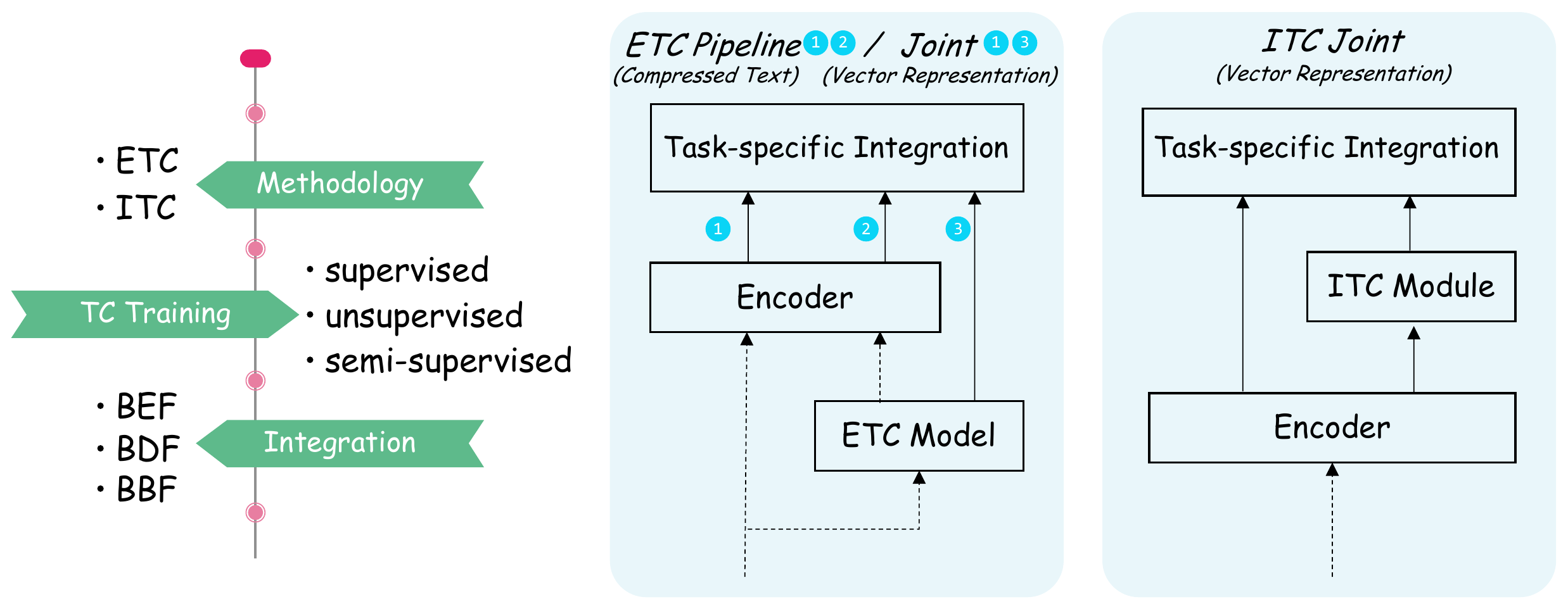}
		\caption{\textcolor{black}{An overview of our text compression settings (left) and overall architectures of ETC and ITC (right). In the right sub-figures, the solid arrows represent differentiable operations and the dashed represent non-differentiable operations. \textit{ETC Pipeline}, \textit{ETC Joint}, and \textit{ITC Joint} are three manners explored for text compression-aided Transformer encoding in this paper. In \textit{ETC Pipeline} and \textit{ETC Joint} manners, text is compressed using an independent model, denoted as ETC Model. In the \textit{ITC Joint} manner, text compression is handled in the full model in a module we denote ITC Module.}}
		\label{fig:overview}
	\end{figure}

	\textcolor{black}{Our approach aims to integrate the products of compression (text sequence or feature vectors) into the encoding of vanilla Transformer to improve the resulting representations.
	We classify two types of text compression, Explicit Text Compression (ETC) and Implicit Text Compression (ITC), which signify whether the compressed sequence is explicitly generated before being passed to the downstream model or implicitly generated as feature vectors by a part of the downstream model, respectively. 
    In Figure \ref{fig:overview}, we show an overview of our approach and the overall architectures of ETC and ITC.}
    In our framework, there are two models/modules: compression model (ETC/ITC) and downstream task models. We will first introduce two types of compression models (modules) and then show how to incorporate these compression models (modules) into task models.
	
    \textcolor{black}{
    For ETC, since compression is handled by an external model, we have two options: use a pipeline manner, whereby a text compression model is trained before training the downstream model (\textit{ETC Pipeline}) or train the text compression model and the downstream task model jointly (\textit{ETC Joint}). 
    For ITC, since the compressed features is directly generated by a module inside the downstream task's model, jointly training the downstream task model with the objective of text compression is the only solution. We thus call this method \textit{ITC Joint}.
    Thus, we have three text compression-aided model training manners: \textit{ETC Pipeline}, \textit{ETC Joint}, and \textit{ITC Joint}. 
    }
	
	\textcolor{black}{Depending on whether or not the text compression training requires \textbf{human annotated data}, the text compression can be trained in three settings: {supervised}, {unsupervised}, and {semi-supervised}. 
	Generally, downstream models can be roughly divided into two types: \textit{encoder-decoder} and \textit{encoder-classifier}.
	For the downstream models, we propose different methods of integrating the compressed text and fusing its representation with that of the original Transformer: Backbone Encoder-side Fusion (\textbf{BEF}), Backbone Decoder-side Fusion (\textbf{BDF}), and Backbone Both-side Fusion (\textbf{BBF}).}

	\section{Text Compression}
	
	\subsection{Explicit Text Compression}\label{sec:etc}
	
	Generally, text summarization is a typical sequence generation task that aims to maximize the absorption and long-term retention of large amounts of data over a relatively short sequence for text understanding~\cite{knight2002summarization,che2015sentence}\footnote{Text summarization typically consists of two types: extractive and abstractive. The extractive type has been well studied. We focus on abstractive summarization in this paper.}. 
	To distinguish the importance of words in a sentence or paragraph and, more importantly, to dig out the most salient parts of the source representation and emphasize these parts, we use text summarization, and term this task \textbf{ETC} in this paper. 
	
	ETC can be conducted by a typical sequence-to-sequence model. The encoder represents the input sentence $x$ as a sequence of vectors, and the autoregressive decoder uses the attention mechanism to learn the context vector for generating a text compressed sequence $x_c$, which contains the key meaning of the input sequence.
	Recently, the new Transformer architecture proposed by Vaswani et al. \cite{vaswani2017attention}, which fully relies on self-attention networks, has exhibited state-of-the-art translation performance for several language pairs. We attempt to apply the Transformer architecture to such a compression task. 
	
	\textcolor{black}{When applying ETC to downstream tasks, the ETC model and downstream task model are optimized separately, forming a pipeline manner, and we refer to this manner as \textit{ETC Pipeline}. For the downstream task, the ETC Pipeline is simple and less invasive to the downstream model; however, the outputs of the low-level ETC model produce errors that can propagate to the downstream task's model and hamper performance.}
	
	\textcolor{black}{Multi-Task Joint Learning (MTL) is popular learning paradigm and promising technique in machine learning that aims to leverage useful information contained in multiple related tasks to help improve the generalization performance of all the tasks. Since both the downstream task and text compression can benefit from being performed jointly, it is very intuitive to train the ETC model with the downstream model jointly, namely \textit{ETC Joint}. In \textit{ETC Joint}, the Transformer output vectors in the decoder are directly input to the downstream model as a compressed text feature instead of re-encoding the decoded compressed text sequence.} 
	
	\noindent\textbf{ETC Compression Rate Control} Explicit compression rate (length) control is a common method that has been used in previous text compression works. Kikuchi et al. \cite{kikuchi-etal-2016-controlling} examined several methods of introducing target output length information and found that they were effective without negatively impacting summarization quality. Fan et al. \cite{fan-etal-2018-controllable} introduced a length marker token that induces the model to target an output of a desired length that is coarsely divided into discrete bins. Fevry and Phang \cite{fevry-phang-2018-unsupervised} augmented the decoder with an additional length countdown input, which is a single scalar that ticks down to $0$ when the generation reaches the desired length.
	
	In \textit{ETC Pipeline}, different from the length marker or length countdown input, to induce our ETC model to output a compressed sequence with a desired length, we use beam search during generation to find the sequence $x_c$ that maximizes a score function $s(x_c, x)$ given a trained ETC model. The length normalization is introduced to account for the fact that we have to compare hypotheses of different lengths. Without some form of length-normalization \textit{LenNorm}, beam search will favor shorter sequences over longer ones on average since a negative log-probability is added at each step, yielding lower (more negative) scores for longer sentences. Moreover, a coverage penalty $cp$ is also added to favor the sequence that covers the source sentence meaning as much as possible according to the attention weights \cite{wu2016google}.
	\begin{align}
	s(x_c, x) = log(P(x_c|x)) &/ \textit{LenNorm}(x_c) + cp(x; x_c),\\
	\textit{LenNorm}(x_c) = &(5 + |x_c|)^{\alpha} / (5 + 1)^{\alpha}, \\
	cp(x; x_c) = \beta \times &\sum_{i=1}^{|x|} log(min(\sum_{j=1}^{|x_c|}p_{i,j}, 1.0)),
	\end{align}
	where $p_{i,j}$ is the attention probability of the $j$-th target word on the $i$-th source word. Parameters $\alpha$ and $\beta$ control the strength of the length normalization and the coverage penalty. Although $\alpha$ can be used to control the compression ratio softly, we use the compression ratio $\gamma$ to control the maximum length of decoding generation by hard requirements. When the decoding length $|x_c|$ is greater than $\gamma |x|$, decoding stops.
	
	\textcolor{black}{In \textit{ETC Joint}, for the feasibility of mini-batch training, greedy decoding is performed based on the maximum compression length in a batch, and then the compressed representations are masked according to their respective desired lengths to ensure that the compression rate is correctly controlled.}

	\begin{figure}
		\centering
		\includegraphics[width=0.5\textwidth]{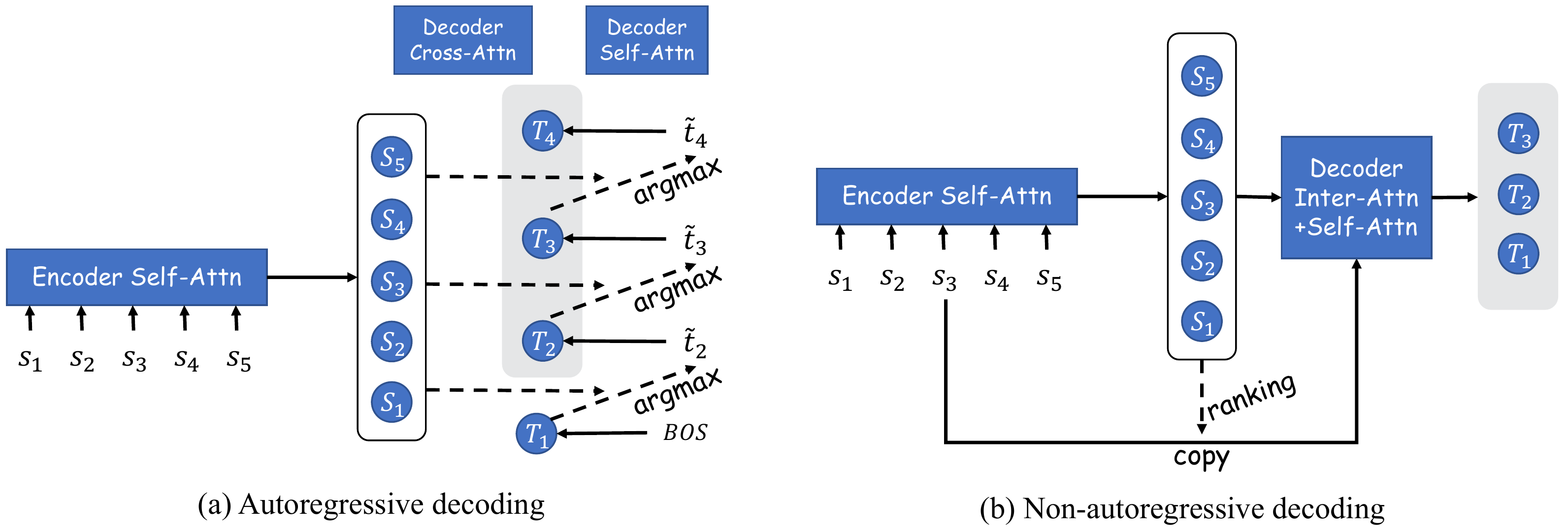}
		\caption{\textcolor{black}{Comparison between autoregressive decoding in \textit{ETC Joint} and non-autoregressive decoding in \textit{ITC Joint}. The solid line represents differentiable operations, and the dashed line represents non-differentiable ones. The gray background represents the compressed text features. $s$ is used for tokens in the source side, $\hat{t}$ for predicted tokens in the target side, and $BOS$ for tokens to start the prediction.}}
		\label{fig:at_nat}
	\end{figure}
	
	\begin{figure*}
		\centering
		\includegraphics[width=0.8\textwidth]{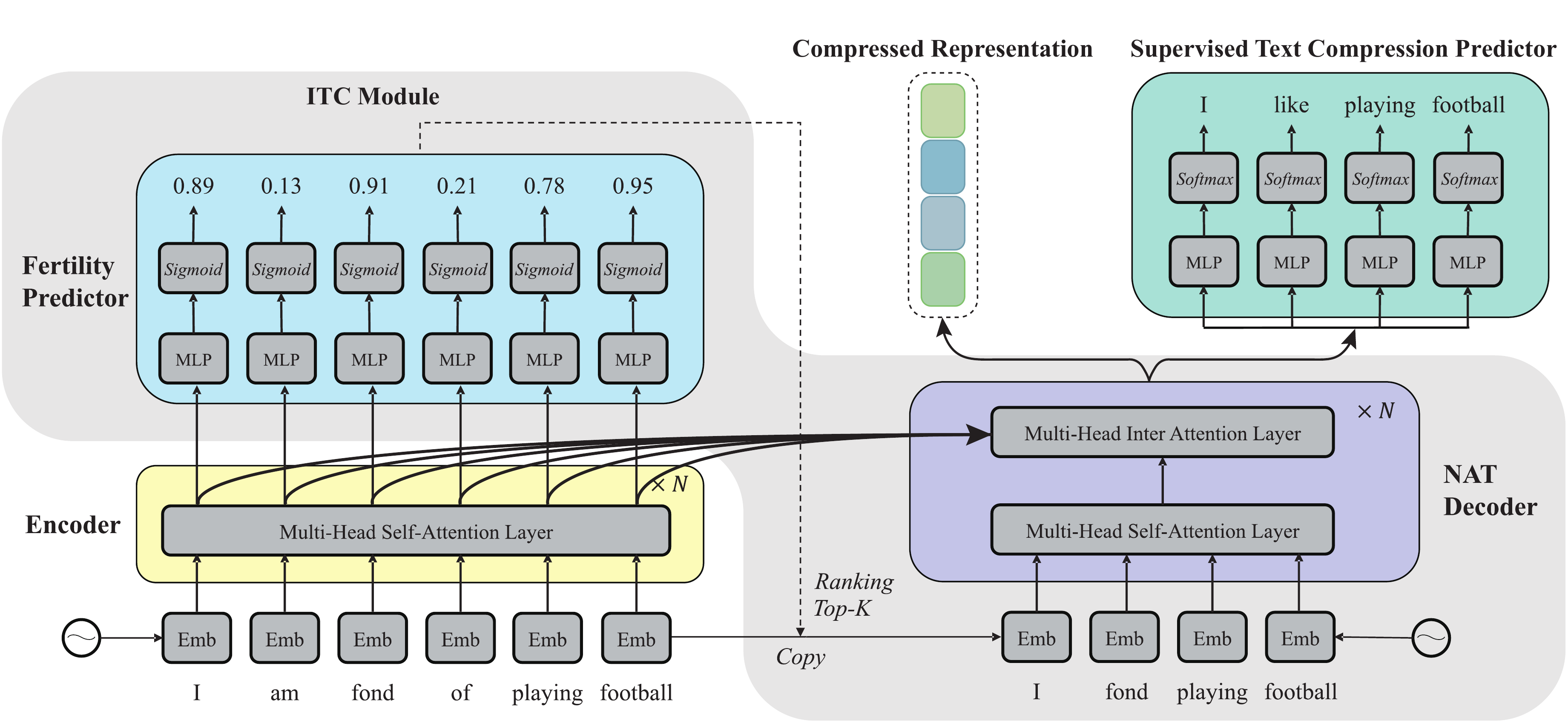}
		\caption{The architecture of the ITC module.}
		\label{fig:itc}
	\end{figure*}
	
	\subsection{\textcolor{black}{Implicit Text Compression}}
	
	\textcolor{black}{
	Implicit Text Compression (ITC) is a method that allows text compression to be trained jointly with the downstream task.
	The errors generated by \textit{ETC Pipeline}'s text compression model are not differentiable across the two models used, as these models pass hard text sequences rather than soft representations, 
	and though \textit{ETC Joint}'s gradient is differentiable for the self-attention layer of the ETC decoder, the encoder and encoder-decoder attention of the method's text compression model will not be updated during the autoregressive decoding process, as shown in Figure \ref{fig:at_nat}(a).
	In addition to this lack of differentiability, the left-to-right sequential decoding method of autoregressive decoding means hidden states must be generated one by one and cannot be parallelized, meaning time cost is another consideration.
	To address this issue, ITC uses a non-autoregressive decoder \cite{gu2018nonautoregressive} that allows full integration of text compression into the downstream model and allows for a fully differentiable gradient, as seen in Figure \ref{fig:at_nat}(b). We name this approach \textit{ITC Joint}. With this novel approach, we no longer rely on generating a compressed text sequence (which is needed for \textit{ETC Joint}, as the generation of a subsequent token depends on the previous one). 
	}
	
	\textcolor{black}{Compared with autoregressive decoding, non-autoregressive decoding not only has the advantage of removing non-differentiable operations and thus allowing the whole model to be optimized at the same time, but it has excellent decoding time complexity ($O(1)$ compared to $O(n)$ in autoregressive decoding) that greatly speeds up the joint training and inference of the full model. }
    
    We present the architecture of the ITC module in Figure \ref{fig:itc}. The ITC module consists of three main components: a fertility predictor, a Non-Autoregressive Transformer (NAT) decoder, and an optional text compression predictor.
	
	\noindent\textbf{Fertility Predictor} Unlike the non-autoregressive machine translation model in \cite{gu2018nonautoregressive}, the fertility predictor in the ITC module we proposed is not used to solve multimodality problems in different languages. In machine translation, a word in the source language may correspond to zero or more words in the target language due to the differences between languages. Because of the conditional independence assumption of NAT machine translation, predicting this correspondence relationship before inputting it to the decoder is necessary. Thus, the fertility predictor is designed to predict the number of copies of words, known as the fertility, in the source language. The NAT decoder then uses this prediction to copy each encoder input as a decoder input zero or more times.
	
	In text compression, because the text needs to be compressed and thus shorter, the input word usually does not correspond to a word in the target sentence or only corresponds to one single word. 
	Our fertility predictor is used to predict the probability that a word appears in the target sequence rather than predicting the number of copies as is done in typical fertility predictors. Thus, we adopt a one-layer neural network with a sigmoid function instead of a softmax classifier. It is formulated as follows:
	\begin{equation}
	p_f = \sigma(\textbf{MLP}(H_x)).
	\end{equation}
	
	Since text compression is simpler than machine translation, the order of the compressed text will not change too much. In the text compression process, the model will extract the backbone of the sentence and paraphrase it to produce a shorter sentence. 
	
	\noindent\textbf{NAT Decoder} The input for the NAT decoder is a copy of the encoder input that is compressed to its top-$K$ words according to the fertility sort, where
	\begin{equation}
	K = \gamma|x|.
	\end{equation}
	
	In the NAT decoder, without the constraint of an autoregressive factorization of the output sequence, earlier decoding steps do not risk accessing information in later steps. Thus, we can avoid the causal mask used in the self-attention layer of the conventional Transformer's decoder and use the same multi-head self-attention layer as in the encoder. The compressed representation $H_x^c$ is
	\begin{equation}
	\begin{split}
	H_{itc} &= \textup{FFN}(\textup{MultiHead}(\textbf{Q}, \textbf{K}, \textbf{V}))+\textbf{V}, \\
	H_x^c &= \textup{FFN}(\textup{MultiHead}(H_{itc}, H_x, H_x)) + H_{itc}.
	\end{split}
	\end{equation}
	
	This is unlike the original NAT Decoder implementation in \cite{gu2018nonautoregressive}, which masks each query position, preventing it from attending to itself. 
	In addition, because the word order of compressed text does not change much, the additional positional attention mechanism is not included.
	
	\noindent\textbf{Optional Text Compression Predictor} For ITC, compressed representations can be obtained using only the fertility predictor and the NAT decoder, which are optimized along with the downstream task models. To expose the ITC module to true text compression signals for pre-training, we added an optional compression predictor to output compressed text sequence distributions and enabled straightforward maximum likelihood training with a text compression target. More precisely, given a source sentence $x$, the conditional probability of a target compression sequence $y^c$ is:
	
	\begin{equation}
	p(y^c|x;\theta) = \sum_{f_i \in \mathcal{F}} (p_f(f_i)\prod_{i=1}^{K}p_c(y^c_i|x^{\{f_i\}}_1, ..., x^{\{f_i\}}_K)); \theta)),
	\end{equation}
	where $\mathcal{F}$ is the set of all fertility sequences, and $x^{\{f_i\}}$ denotes tokens in the top-$K$ filtered sequence $f_i$.
	
	\section{Text Compression Training}
	
	Typically, downstream task can be done on several levels, including on an individual sentence level or paragraph level or on the entire document as a whole. To adapt text compression to various downstream tasks, we train text compression at the sentence and paragraph levels. These two types of models are only distinguished by the maximum input length limit and the training datasets.
	
	\noindent\textbf{Supervised Training} Text compression usually relies on large-scale raw data together with human-labeled data, which acts as supervision, to train a compression model~\cite{rush-etal-2015-neural,hu-etal-2015-lcsts,chopra-etal-2016-abstractive,cheng-lapata-2016-neural,nallapati-etal-2016-abstractive,duan-etal-2019-zero}.
	
	For supervised training, we used the Annotated Gigaword corpus~\cite{napoles-etal-2012-annotated} as the sentence-level training dataset and CNN/DailyMail (CNN/DM) \cite{hermann2015teaching} as the paragraph-level training dataset.
	The Gigaword dataset is derived from news articles and consists of pairs of the main sentences in the article (longer) and the headline (shorter). It includes approximately 3.8M training samples, 400K validation samples, and 2K test samples. The CNN/DM dataset is collected from the CNN and the Daily Mail news websites. Both news providers supplement their articles with a number of bullet points, summarizing aspects of the information contained in the article. These summary points are abstractive and do not simply copy sentences from the documents. 
	
	\noindent\textbf{Unsupervised Training} A major challenge in supervised text compression training is the scarcity of high-quality human annotated parallel data. In practice, supervised training often cannot be done due to a lack of annotated data. In particular, models also often suffer from poor domain adaptation as the domain variety of the available data is also insufficient. 
	Therefore, the effectiveness of models relies heavily on the availability of large amounts of parallel original sentences and human-annotated compressed sentences. This hinders text compression from further improvements for many low-resource scenarios. 
	
	Studies of human summarizers show that it is common to apply various other operations while compressing, such as paraphrasing, generalization, and reordering \cite{jing-2002-using}. 
	Fevry and Phang \cite{fevry-phang-2018-unsupervised} added noise to extend the original sentences to train the text compression model. Adding noise creates parallel pseudo data, which is used for unsupervised training.
	In this setup, the model has to exclude and reorder the noisy sentence input and hence learn to output more semantically important and shorter but grammatically correct sentences. 
	Inspired by this, we use three types of noise for data synthesis: \textbf{Additive Sampling Noise}, \textbf{Shuffle Noise}, and \textbf{Word Dropout Noise}.
	
	\begin{itemize}
		\item \textbf{Additive Sampling Noise:} To extend the original input, we sample additional instances from the training dataset randomly and then sub-sample a subset of words from each of these instances without replacement. The newly sampled words are appended to the end of the original sequence.
		\item \textbf{Shuffle Noise:} In the sentence-level model, to encourage the model to learn to rephrase the input sequence to make the output shorter, we shuffle the resultant additive noisy sequence. In the paragraph-level model, a paragraph is divided into sentences based on full stops, and these sentences, together with the sampling sequence, are then shuffled in random order.
		\item \textbf{Word Dropout Noise:} In addition to the above two noises introduced by Fevry and Phang \cite{fevry-phang-2018-unsupervised}, we introduce word dropout noise to enable the model to predict tokens unseen in the input, which leads to abstract outputs. 
	\end{itemize}
	
	\noindent\textbf{Semi-supervised Training} As noted in Song et al. \cite{song2019mass}, the sequence-to-sequence framework has attracted much attention recently due to the advances of deep learning using large-scale data. Many language generation tasks have only a small amount of paired data, which is insufficient for training a deep model with good generalization ability. In comparison, there is a lot of unpaired data, which is easier to obtain.
	
	We observe difference performance in different domains in the supervised training. According to the experimental results of Fevry and Phang \cite{fevry-phang-2018-unsupervised}, the accuracy of unsupervised training is currently lower than that of supervised training. Therefore, we adopt semi-supervised training to alleviate this problem. Specifically, unsupervised training (often referred to as pre-training) is performed using the unpaired data first before finetuning with the small amount of paired data (supervised training) to obtain a model with good performance and generalization ability. 

	\section{Text Compression Integration}
	
	In this section, we introduce the general downstream models and the methods of integrating compressed text in in our approach. To meet the diverse needs of downstream models, based on the fusion position of the backbone information, we propose three novel methods of integrating compressed text: Backbone Encoder-side Fusion (BEF), Backbone Decoder-side Fusion (BDF), and Backbone Both-side Fusion (BBF). 
	
	\subsection{Downstream Model Paradigm}
	
	First, we introduce two commonly used downstream model paradigms: \textit{encoder-decoder} and \textit{encoder-classifier}.
	
	\noindent\textbf{Encoder-Decoder} A Transformer NMT model is a typical \textit{encoder-decoder} downstream model which consists of an encoder and a decoder. It fully relies on self-attention networks (\textbf{SANs}) to translate a sentence in one language to another language while maintaining meaning.
	
	The encoder is described in section \ref{encoder}. Here, we focus on the decoder.
	The SAN of the decoder uses both encoder output $H_x$ and target context hidden state $H_{tgt}$ to learn the context vector $o_i$ by ``encoder-decoder cross-attention":
	\begin{equation}
	\begin{split}
	H_{tgt} &= \textup{FFN}(\textup{MultiHead}_{masked}(\textbf{Q}, \textbf{K}, \textbf{V}))+\textbf{V}, \\
	c_i &= \textup{FFN}(\textup{MultiHead}(H_{tgt}, H_x, H_x)),
	\end{split}
	\label{eq4:context_self-attention}
	\end{equation}
	\begin{equation}
	o_i = c_i + H_{tgt}.
	\label{eq:tgt_fusion}
	\end{equation}
	
	Then, the context vector $o_{i}$ is used to compute translation probabilities of the next target word $\textit{y}_i$ by a linear, potentially multi-layered function:
	\begin{equation}
	P(\textit{y}_i | \textit{y}_{<i}, \textit{x}) \propto 
	\textup{Softmax}(\textbf{L}_\textit{o} \textbf{GeLU}(\textbf{L}_\textit{w}o_{\textit{i}})),
	\label{eq5:Probabilities}
	\end{equation}
	where $\textbf{L}_{o}$ and $\textbf{L}_{w}$ are projection matrices.
	
	\noindent\textbf{Encoder-Classifier}
	The \textit{encoder-decoder} is a very general model paradigm, but most of the tasks in NLP follow the paradigm of \textit{encoder-classifier}. 
	We take two popular styles of MRC tasks, span-based and multi-choice, as examples for illustration.
	For the span-based style, the query is a question whose answers are spans of texts. 
	For the multi-choice style, the model is requested to choose the right answer from a set of candidate ones according to given passages and questions. 
	
	Formally, we describe the reading comprehension task as a tuple $<P, Q, (O), A>$, where $P$ is a passage (context), $Q$ is a query over the contents of $P$, and a span or option $O$ is the right answer $A$. Given the original passage $P$ and a question $Q$, we organize its input $x$ for the encoder as the following sequence:
	\begin{align*}
	\textbf{Span:} & \quad [CLS] \quad P \quad [SEP] \quad Q \quad [SEP], \\
	\textbf{Choice:} & \quad [CLS] \quad P || Q \quad [SEP] \quad O \quad [SEP].
	\end{align*}
	where ``$||$" denotes concatenation operation, and both of $[CLS]$ and $[SEP]$ denote special marks.
	
	The sequence (Span or Choice) is then fed to the encoder to attain the contextualized representation $H_x$. In the span-based style model, $H_x$ is fed to answer position network~\cite{vinyals2015pointer} to compute the start position $A_s$ and end position $A_e$ for the answer span:
	\begin{equation}
	\begin{split}
	&  A_s = \textup{Ptr-Net}_s(H_x),   A_e = \textup{Ptr-Net}_e(H_x). 
	\end{split}
	\label{eq14:start_end_pos}
	\end{equation}
	
	For the tasks such as SQuAD 2.0 challenge \cite{rajpurkar-etal-2018-know} that contains unanswerable questions, we define an answer of 0 for both the start and end positions as a prediction of it being unanswerable, or we pass $H_x$ to one additional verifier module to predict whether the question is answerable or not.
	
	In the multi-choice style, $H_x$ is fed to one MLP classifier to predict the choice label $A_c$:
	\begin{equation}
	P(A_c) \propto \textup{Softmax}(\textup{MLP}(H_x)).
	\label{eq15:choice_label}
	\end{equation}
	
	\begin{figure*}[thb!]
		%\centering
		\subfigure{
			\begin{minipage}[b]{0.48\linewidth}
				\centering
				\includegraphics[width=0.98\textwidth]{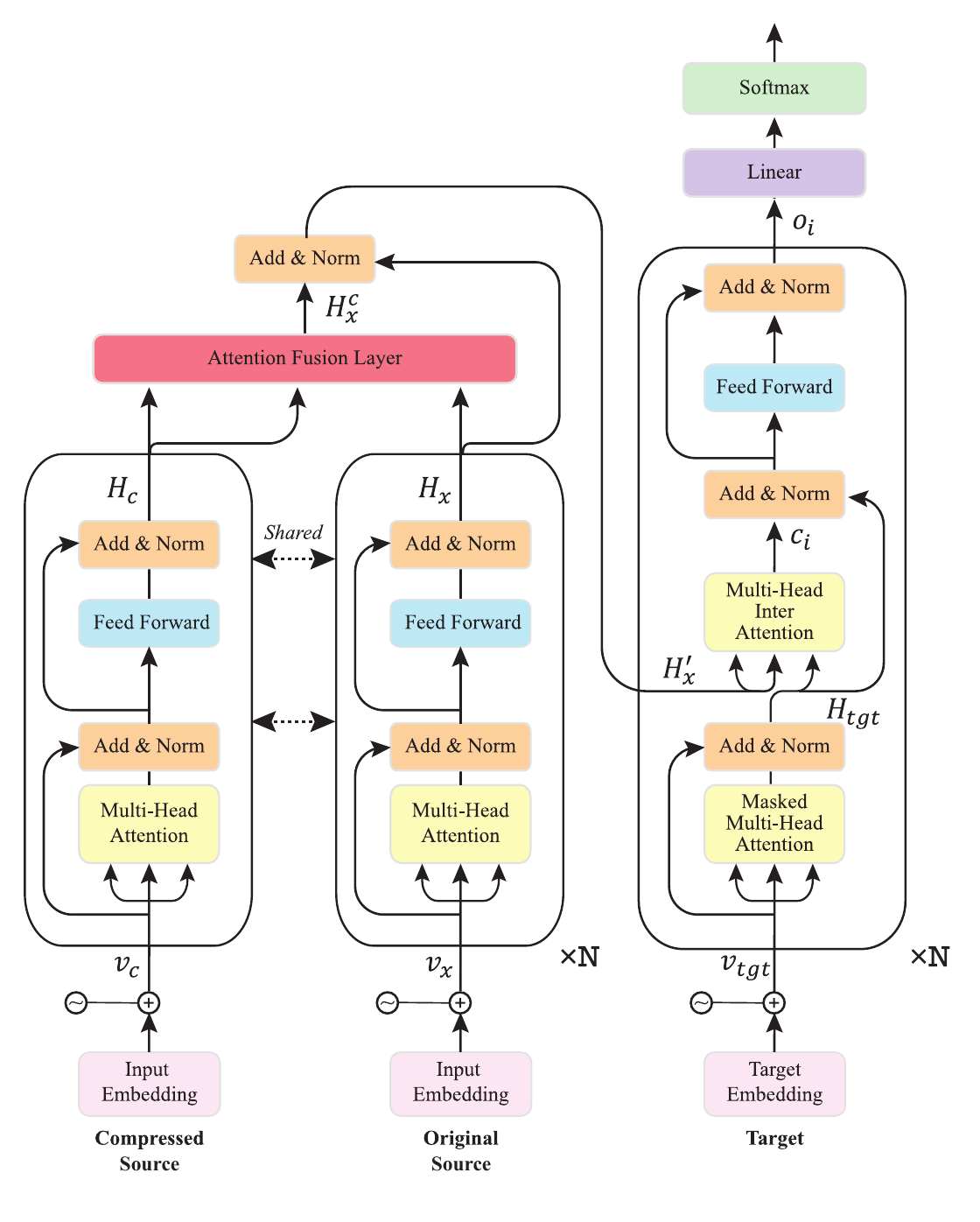}
				\caption{The architecture of our NMT model with \textbf{\textit{ETC Pipeline} BEF}.}
				\label{fig:befnmt}
			\end{minipage}
		}
		\subfigure{
			\begin{minipage}[b]{0.48\linewidth}
				\centering
				\includegraphics[width=0.98\textwidth]{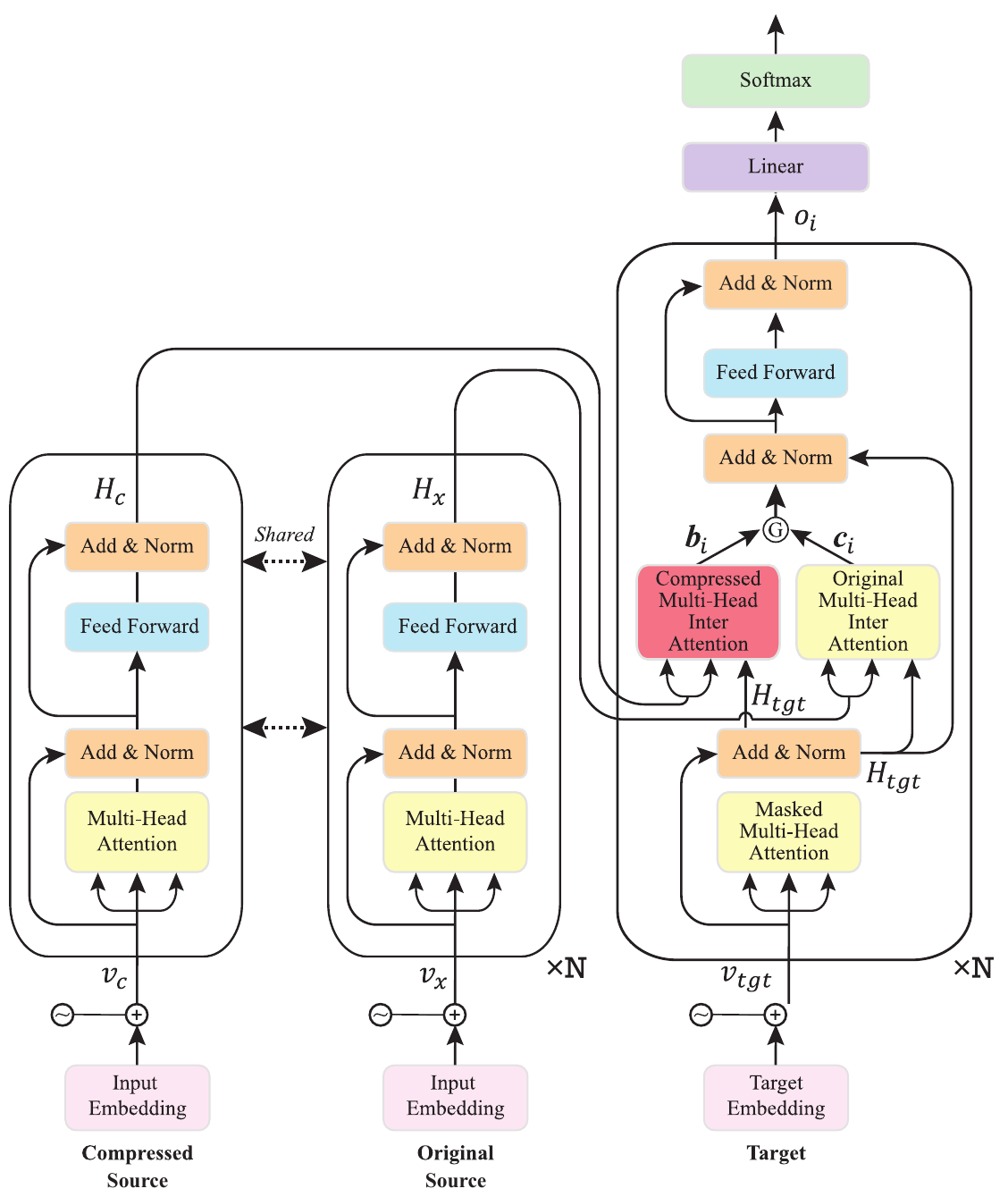}
				\caption{The architecture of our NMT model with \textbf{\textit{ETC Pipeline} BDF}.}
				\label{fig:bdfnmt}
			\end{minipage}
		}
	\end{figure*}

	\subsection{Backbone Encoder-side Fusion}\label{bsrlf}
	
	\textcolor{black}{In BEF, the backbone sequence (in \textit{ETC Pipeline}) or hidden states  (in \textit{ETC Joint} and \textit{ITC Joint}) are integrated with the original Transformer representations in the encoder side. Figure \ref{fig:befnmt} shows BEF on the Transformer NMT model with \textit{ETC Pipeline}.}
	In \textit{ETC Pipeline}, given an input sequence $x$=$\{x_1, \cdots, x_J\}$, there is an additional compressed sequence $x_c$=$\{x^c_1, \cdots, x^c_K\}$ of length $K$ generated by the proposed ETC model.
	This compressed sequence is also input to the SAN shared with the original encoder with word vectors $v_c = \{v^c_1, \cdots, v^c_K\}$ in a shared vocabulary to learn the compressed sequence's final representation $H_{c}$=$\{H^{c}_1, \cdots, H^{c}_K\}$.
	\textcolor{black}{In \textit{ETC Joint} and \textit{ITC Joint}, the hidden states in text compression decoder are directly used as the final representation $H_{c}$ of the compressed sequence.}
	We introduce an additional multi-head attention layer to fuse the compressed representation and the original Transformer encoding to learn a more effective representation.
	
	For the multi-head attention-fusion layer, a compressed text representation $H_x^c$ is computed by the multi-head attention on the original sequence representation $H_x$ and the compressed sequence representation $H_c$:
	\begin{equation}
	\begin{split}
	H_x^c = \textup{FFN}(\textup{MultiHead}(H_x, H_c, H_c)).
	\end{split}
	\end{equation}
	
	$H_x^c$ and $H_x$ are added to form a fusion source representation $H_{x}^{'}$: 
	\begin{equation}
	H_{x}^{'} = H_x +H_x^c.
	\label{Eq9:fuse_src_rep}
	\end{equation}
	
	Finally, the $H_{x}^{'}$ (instead of $H_{x}$) is input to Eq.~\eqref{eq4:context_self-attention} for predicting the target translations or Eq.~\eqref{eq14:start_end_pos} for predicting the start position $A_s$ and end position $A_e$ for the answer span.
	Similarly, the $H_x^{'}$ is also input to Eq.~\eqref{eq15:choice_label} for predicting the choice label $A_c$.

	\subsection{Backbone Decoder-side Fusion}
	
	\textcolor{black}{BEF is applicable to both \textit{encoder-decoder} and \textit{encoder-classifier} downstream models as long as a encoder is used; however, because the \textit{encoder-decoder} paradigm uses cross-attention (``encoder-decoder inter-attention") to select the encoder's representation (i.e. source representation), we propose an additional method of integration. In this method, the compressed representation is regarded as a new independent source representation, and the decoder will handles the two representations separately. We call this BDF. We present BDF with the Transformer NMT model and \textit{ETC Pipeline} in Figure \ref{fig:befnmt}.}
	
	In BDF, the original Transformer encoding and the compressed representation are represented as $H_x$ and $H_c$ respectively.
	We then use a tuple ($H_x,H_c$) instead of the encoder-side fusion representation $H_x^{'}$ as the input to the decoder.
	Specifically, we introduce an additional ``encoder-decoder inter-attention" module into the decoder to learn the compressed sequence context $b_i$ at the current time-step $i$:
	\begin{equation}
	b_i=\textup{FFN}(\textup{MultiHead}(H_{tgt}, H_c, H_c)).
	\label{Eq11:skeleton_ctx}
	\end{equation}
	
	Since our goal is to treat the original encoding and the compressed representation as two independent contexts, we use a context gate $g_c$ for integrating two independent contexts: original context $c_i$ and compression context $b_i$. The gate $g_i$ is calculated by:
	\begin{equation}
	g_i = \sigma(\textbf{MLP}([c_i; b_i])).
	\end{equation}
	
	Therefore, the final decoder-side fusion context $c_i'$ is:
	\begin{equation}
	c_i' = g_i \otimes c_i + (1 - g_i) \otimes b_i,
	\end{equation}
	where $\sigma$ is the logistic sigmoid function, $\otimes$ is the point-wise multiplication, and $[\cdot]$ represent the concatenation operation.
	
	The context $c_i'$ is input to replace the $c_i$ the Eq.~\eqref{eq:tgt_fusion} to compute the probabilities of next target prediction.
	
	\subsection{Backbone Both-side Fusion}
	
	\textcolor{black}{As discussed in the BEF and BDF, the compressed representation in \textit{encoder-decoder} paradigm can be used both to improve source-side encoding and to provide additional representation for the decoder, so a novel BBF can be obtained by combining these two methods of integration.}

	In BBF, both the original representation $H_x$ and text compression-aided representation $H_x^{'}$ are input to the decoder. Similarly, we introduce an additional ``encoder-decoder inter-attention" module into the decoder to learn a text compression-aided context $b_i^{'}$ at the current time-step $i$:
	\begin{equation}
	b_i^{'}=\textup{FFN}(\textup{MultiHead}(H_{tgt}, H_{x}^{'}, H_{x}^{'})).
	\end{equation}
	
	Then, the context gate $g_i$ in BBF (consistent with BDF) is applied to combine the two contexts $c_i$ and $b_i^{'}$.

	\section{Experimental Setup}\label{sec:setup}
	
	\subsection{Text Compression} 
	
	\textcolor{black}{In ETC, under the {supervised} setting, we use the Transformer (big) architecture \cite{vaswani2017attention}. The number of layers/hidden size/FFN size/number of heads is 6/1024/4096/16, and only the text compression objective is used for model training (without any pre-training approaches). Under the {unsupervised} and {semi-supervised} settings, in order to make full use of the existing pre-trained resources, we use the same architecture as BART (large) \cite{lewis2019bart} with number of layers/hidden size/FFN size/number of heads 12/1024/4096/16, and the model is initialized with a pre-trained BART checkpoint before training.}
	
	\textcolor{black}{In ITC, since the ITC module is part of the downstream model and the encoder is shared, we only need to add an additional NAT decoder. The number of layers/hidden size/FFN size/number of heads is 6/512/2048/8. An additional linear mapping layer is added to handle the inconsistency of hidden size with the downstream model. The ITC-aided model is two-stage trained. The first stage uses text compression data for pre-training, and the second stage uses downstream task targets to train the model as a whole. Under the {supervised}/{unsupervised}/{semi-supervised} settings, only the text compression training data used in the first stage is different.}
	
	\textcolor{black}{For the text compression training data, we use Gigaword for sentence-level and CNN/DM for paragraph-level in {supervised} training, use a pseudo-parallel corpus synthesized from 190M English monolingual unpaired data from the WMT News Crawl datasets in the {unsupervised} setting (same as ~\cite{song2019mass}), and combine the real and synthesized text compression data for {semi-supervised} model training.}
	
	To evaluate the quality of text compression, the $F_1$ score of ROUGE-1 (R-1), ROUGE-2 (R-2), and ROUGE-L (R-L)~\cite{lin-2004-rouge} was used to evaluate the model. In \textit{ETC Pipeline}, we used beam search with a beam size of 5, a length normalization of 0.5, and a coverage penalty of 0.2. 
	Baseline systems in the sentence-level text compression include AllText and F8W \cite{rush-etal-2015-neural,wang-lee-2018-learning}. 
	F8W is simply the first 8 words of the input, and AllText uses the whole text as the compression output. We also present the Lead-3 baseline, which simply selects the first
	three sentences for a given input in the paragraph-level text compression.
	We also included the following other pre-training methods: masked language modeling (MLM, BERT) \cite{devlin-etal-2019-bert}, denoising auto-encoder (DAE) \cite{vincent2008extracting}, and masked sequence to sequence (MASS) \cite{song2019mass} to compare with our unsupervised pre-training method in the semi-supervised setting. \textcolor{black}{Since \textit{ETC Joint} can not be fully trained and was slow in training, we did not include this in the main experiments. Only \textit{ETC Pipeline} is used in the main experiment and \textit{ETC Joint} is left for analysis and comparison in the ablation experiment. Because ITC does not exist as an independent model, the model obtained from the first stage training of the ITC-aided Transformer-base model in the following WMT14 EN-DE machine translation experiment was selected as the sentence-level evaluation object, and ITC-aided RACE reading comprehension model (BERT$_{base}$ baseline) was selected as the paragraph-level evaluation object.}
	
	\textcolor{black}{In sentence-level ETC, we set the compression rate $\gamma$ to 0.6, and in paragraph-level ETC, we set the compression rate to 0.3. The compression rates of sentence-level and paragraph-level ITC are both set to 0.4.}

	\subsection{Machine Translation} 
	
	The proposed NMT model was evaluated on the WMT14 English-to-German (EN-DE) and English-to-French (EN-FR) tasks, which are both standard large-scale corpora for NMT evaluation.
	For the EN-DE translation task, 4.43M bilingual sentence pairs from the WMT14 shared task, which includes Common Crawl, News Commentary, and Europarl v7, were used as training data.
	The newstest2013 and newstest2014 sets were used as the dev set and test set, respectively.
	For the EN-FR translation task, 36M bilingual sentence pairs from the WMT14 shared task were used as training data.
	The newstest12 and newstest13 were combined for validation, and the newstest14 was the test set, following the setting of \cite{gehring2017convolutional}. 
	The BPE algorithm~\cite{sennrich-etal-2016-neural} was also adopted, and the joint vocabulary size was set at 40K.
	For the hyper-parameters of our Transformer (base/large) models, we followed the settings used in  \cite{vaswani2017attention}.
	
	In addition, we also reported the state-of-the-art results in recent literature, including modeling local dependencies (\textbf{Localness})~\cite{yang-etal-2018-modeling}, fusing multiple-layer representations in SANs (\textbf{Context-Aware})~\cite{yang2019context}, and fusing all global context representations in SANs (\textbf{global-deep context})~\cite{dou-etal-2018-exploiting}.
	MultiBLEU was used to evaluate the translation task.
	Since the machine translation task we used is a sentence level task, we used sentence-level text compression for enhancement. \textcolor{black}{In addition, for the NMT model with the \textit{encoder-decoder} paradigm, we empirically demonstrated that BBF is a more effective method of integration for \textit{ETC Pipeline}; therefore, on \textit{ITC Joint}, only the results of BBF are reported to save space.}
	
	\begin{table}[h]
		\centering
		\caption{\textcolor{black}{ROUGE performance of the sentence-level text compression models trained on the Gigaword dataset.}}\label{tab:sc_results}
		\begin{tabular}{lccc}
			\toprule
			\textbf{Model}& \textbf{R-1}& \textbf{R-2} & \textbf{R-L} \\
			\midrule
			\textit{Baselines:} & & & \\
			All text & 28.91 & 10.22 & 25.08\\
			F8W & 26.90 & 9.65 & 25.19 \\
			\midrule
			\textit{Unsupervised:} & & & \\
			Fevry et al.~\cite{fevry-phang-2018-unsupervised} & 28.42 & 7.82 & 24.95 \\
			\textbf{ETC} & 31.37 & 8.25 & 28.01 \\
			\textbf{ITC} & 30.42 & 8.06 & 27.44 \\
			\midrule
			\textit{Supervised:} & & & \\
			RNN-based Seq2seq & 35.50 & 15.54 & 32.45 \\
			Nallapati et al.\cite{nallapati-etal-2016-abstractive} & 34.97 & 17.17 & 32.70 \\
			\textbf{ETC} & 37.53 & 18.48 & 34.79 \\
			\textbf{ITC} & 36.95 & 17.32 & 34.01 \\
			\midrule
			\textit{Semi-supervised:} & & & \\
			MLM Pre-training & 37.75 & 18.45 & 34.85 \\
			DAE Pre-training & 35.97 & 17.17 & 33.14 \\
			MASS\cite{song2019mass} & 38.73 & 19.71 & 35.96 \\
			\textbf{ETC} & 39.54 & 20.35 & 36.79 \\
			\textbf{ITC} & 38.45 & 19.12 & 35.63 \\
			\bottomrule
		\end{tabular}
	\end{table}

	\subsection{Machine Reading Comprehension}
	
	\noindent\textbf{Span-based MRC}  As a widely used MRC benchmark dataset, SQuAD 2.0 \cite{rajpurkar-etal-2018-know} extends the 100,000 questions in SQuAD 1.1 \cite{rajpurkar-etal-2016-squad} with over 50,000 new, unanswerable questions that are written adversarially by crowd-sourced workers to resemble answerable questions. 
	For the SQuAD 2.0 challenge, systems must not only answer questions when possible but also abstain from answering when no answer is supported by the paragraph, making SQuAD 2.0 an even more challenging natural language understanding task for models.  
	Two official metrics are selected to evaluate model performance: Exact Match (EM) and a softer metric, F1 score, which measures a weighted average of the precision and recall rate at the character level.
	
	\noindent\textbf{Multi-choice MRC} Our Multi-choice MRC is evaluated on the Large-scale ReAding Comprehension Dataset From Examinations (RACE) dataset \cite{lai-etal-2017-race}, which consists of two subsets, RACE-M and RACE-H, corresponding to middle school and high school difficulty levels. 
	RACE contains 27,933 passages and 97,687 questions in total and is recognized as one of the largest and most difficult datasets in multi-choice reading comprehension. The evaluation metric is accuracy.
	
	Since the two MRC tasks are both paragraph-level, paragraph-level text compression and BEF are adopted.

	\begin{table}[h]
		\centering
		\caption{\textcolor{black}{ROUGE performance of the paragraph-level text compression models on the CNN/DM dataset.}}\label{tab:sc_results2}
		\begin{tabular}{l c c c}
			\toprule
			\textbf{Model}& \textbf{R-1}& \textbf{R-2} & \textbf{R-L} \\
			\midrule
			\textit{Baselines:} & & & \\
			Lead-3 & 40.34 & 17.70 & 36.57\\
			\midrule
			\textit{Unsupervised:} & & & \\
			\textbf{ETC} & 38.95 & 17.04 & 35.70 \\
			\textbf{ITC} & 38.77 & 16.25 & 34.13 \\
			\midrule
			\textit{Supervised:} & & & \\
			PTGEN\cite{see-etal-2017-get} & 36.44 & 15.66 & 33.42 \\
			PTGEN+COV\cite{see-etal-2017-get} &  39.53 & 17.28 & 36.38 \\
			TransformerABS\cite{liu-lapata-2019-text} & 40.21 & 17.76 & 37.09  \\
			\textbf{ETC} & 40.45  & 17.92  & 37.14  \\
			\textbf{ITC} & 38.87 & 16.98 & 36.01 \\
			\midrule
			\textit{Semi-supervised:} & & & \\
			UniLM\cite{dong2019unified} & 43.33 & 20.21 & 40.51 \\
			BERTSUMABS\cite{liu-lapata-2019-text} & 41.72 &19.39 & 38.76 \\
			BERTSUMEXTABS\cite{liu-lapata-2019-text} & 42.13 & 19.60 & 39.18 \\
			BART\cite{lewis2019bart} & 44.16 & 21.28 & 40.90\\
			PEGASUS \cite{zhang2019pegasus} & 44.17	& 21.47 &	41.11 \\
			ProphetNet \cite{yan2020prophetnet} & 44.20	& 21.17	& 41.30 \\
			\textbf{ETC} & 44.35  & 21.32  & 41.05  \\
			\textbf{ITC} & 40.74  & 18.19  & 37.25  \\
			\bottomrule
		\end{tabular}
	\end{table}

	\begin{table*}[h]\small
		\centering
		\caption{\label{tab:mt_main_results}Comparison with existing NMT systems on the WMT14 EN-DE and EN-FR translation tasks. ``++/+" after the BLEU score indicates that the proposed method was significantly better than the corresponding baseline Transformer (base or big) at significance level p$<$0.01/0.05\cite{collins-etal-2005-clause}. 	``\#Speed" denotes the decoding speed measured in target tokens per second.}
		\begin{tabular}{llrrlrr}
			\toprule
			\multicolumn{1}{c}{\multirow{1}{*}{System}} &  \multicolumn{1}{c}{\multirow{1}{*}{EN-DE}} & \#Speed & \#Params & \multicolumn{1}{c}{\multirow{1}{*}{EN-FR}} & \#Speed & \#Params   \\ 
			\midrule
			\multicolumn{7}{c}{\textit{Existing NMT systems}}   \\ 
			Transformer (base)~\cite{vaswani2017attention}     & 27.3    & N/A  & 65.0M    & 38.1    & N/A & N/A \\
			\quad+Localness~\cite{yang-etal-2018-modeling} & 28.11 & N/A & 88.8M & N/A & N/A & N/A \\
			\quad+Context-Aware SANs~\cite{yang2019context} & 28.26 & N/A & 194.9M & N/A & N/A & N/A \\
			\quad+global-deep context~\cite{dou-etal-2018-exploiting} & 28.58 & N/A & 111M & N/A & N/A & N/A \\ 
			\hdashline
			Transformer (big)~\cite{vaswani2017attention} & 28.4  & N/A & 213.0M    & 41.0   & N/A & N/A \\ 
			\quad+Localness~\cite{yang-etal-2018-modeling} & 28.89 & N/A & 267.4M & N/A & N/A & N/A \\
			\quad+Context-Aware SANs~\cite{yang2019context} & 28.89 & N/A & 339.6M & N/A & N/A & N/A \\
			\quad+global-deep context~\cite{dou-etal-2018-exploiting} & 29.21 & N/A & 396M & N/A & N/A & N/A \\
			\midrule
			\multicolumn{7}{c}{\textit{Our NMT systems}}  \\ 
			Transformer (base)        &  27.24    & 131K & 66.5M & 38.21    & 130K & 85.7M  \\ 
			\quad\textbf{+\textit{ETC Pipeline} BEF}     &  27.75++    & 121K  & 72.1M  & 39.09++    & 120K & 89.0M \\ 
			\quad\textbf{+\textit{ETC Pipeline} BDF}     &  28.14+    & 120K & 72.7M & 39.22++    & 119K & 89.8M \\ 
			\quad\textbf{+\textit{ETC Pipeline} BBF}    & 28.35++ & 119K & 78.6M & 39.40++ & 116K & 91.4M \\
			\hdashline
			\quad\textbf{+\textit{ITC Joint} BBF}    & 28.57++ & 120K  & 109.2M  & 39.68++  & 116K &  133.5M \\
			\midrule
			Transformer (big)         &  28.23   & 11K & 221.0M  & 41.15    & 11K & 222.3M \\
			\quad\textbf{+\textit{ETC Pipeline} BEF}        &  28.52+   & 10K & 225.2M & 41.92+    & 9K & 227.1M  \\
			\quad\textbf{+\textit{ETC Pipeline} BDF}        &  29.16++  & 9K & 225.7M  & 42.22++    & 8K & 227.5M  \\
			\quad\textbf{+\textit{ETC Pipeline} BBF}     & 29.37++ & 8K & 228.9M & 42.52++ & 8K & 230.3M \\
			\hdashline
			\quad\textbf{+\textit{ITC Joint} BBF}    & 29.60++ & 8K & 273.2M  & 42.77++ & 8K &  275.4M \\
			\bottomrule
		\end{tabular}
	\end{table*}

	\section{Main Results}\label{sec:main_results}
	
	\subsection{Text Compression} 
	
	We conducted a comparison between our proposed text compression model and other text compression models in different settings.

	Table~\ref{tab:sc_results} shows the results at the sentence-level. 
	We observed that the proposed unsupervised ETC model performed substantially better than Fevry and Phang~\cite{fevry-phang-2018-unsupervised}'s unsupervised method.
	The proposed supervised ETC model also substantially outperformed the RNN-based Seq2seq and Nallapati et al. \cite{nallapati-etal-2016-abstractive}'s baseline method; that is, our supervised model gave +2.0 improvements on R-1, R-2, and R-L scores over the RNN-based Seq2seq.
	This means that the proposed Transformer-based approaches can generate compressed sentences of high quality. 
	
	We further compared our semi-supervised model with the semi-supervised pre-training methods of MLM \cite{devlin-etal-2019-bert}, DAE \cite{vincent2008extracting}, and MASS \cite{song2019mass}. Our unsupervised ETC pre-training method consistently outperformed the other unsupervised pre-training ones on the text compression task. \textcolor{black}{Comparing the results of ETC and ITC, because the ITC model is relatively smaller and a pre-trained model such as BART is not used for initialization, the effect is worse. In addition, because the non-autoregressive decoder has some weaknesses in explicit decoding compared to other autoregressive decoders, it only achieves an effect comparable to other explicit autoregressive decoding methods; however, since the strength of ITC lies in the joint training of representations rather than explicit decoding, this is only for rough reference, and the real enhancement is expected to occur on downstream tasks.}
	
	Table \ref{tab:sc_results2} summarizes the results of paragraph-level text compression models on the CNN/DM dataset. First, our unsupervised paragraph-level ETC model does not outperform Lead-3 as it did F8W in the sentence level, suggesting that the paragraph-level text compression task is more complex than at the sentence-level. Second, our supervised paragraph-level ETC model outperforms the TransformerABS reported in \cite{liu-lapata-2019-text}, which indicates that joint training with MLM pre-training objectives can bring some performance improvement during the downstream task target training, which is consistent with the conclusion of 
    work \cite{gururangan-etal-2020-dont}. Third, in the semi-supervised setting, our model is slightly better than our base, BART\cite{lewis2019bart}, in performance, demonstrating the effectiveness of our proposed unsupervised ETC pre-training. Fourth, comparing the unsupervised and supervised results, our unsupervised system outperforms some supervised systems, like PTGEN and PTGEN + COV. This is mainly because our unsupervised model is based on the Transformer structure and initialized with BART's pre-trained parameters, while these supervised systems are built on an RNN structure. This shows that the feature extraction ability of Transformer is stronger than that of RNN and that the BART pre-training is very effective for sequence-to-sequence (seq2seq) generation. This is consistent with results on sentence-level compression. \textcolor{black}{At the paragraph level, especially in the semi-supervised setting, when the goal is a compressed sequence, the advantages of ETC in comparison to ITC are more obvious because of the autoregressive decoding. Since the semi-supervised text compression performs the best, we adopt semi-supervised text compression models for downstream task evaluation when not otherwise specified.}
	
	\subsection{Machine Translation} 
	The main results on the WMT14 EN-DE and EN-FR translation tasks are shown in Table~\ref{tab:mt_main_results}. 
	In the EN-DE task, we made the following observations:
	
	1) The baseline Transformer (base) in this work achieved performance comparable to the original Transformer (base)~\cite{vaswani2017attention}.
	This indicates that it is a strong baseline NMT system.
	
	2) All \textbf{BEF}, \textbf{BDF}, and \textbf{BBF} in \textit{ETC Pipeline} significantly outperformed the baseline Transformer (base/big) and only introduced a very small amount of extra parameters. 
	This indicates that the compressed backbone information was beneficial for the Transformer translation system.
	
	3) \textbf{\textit{ETC Pipeline} BDF} performed better than \textbf{\textit{ETC Pipeline} BEF}. This indicates that the backbone fusion on the decoder-side is better than that on the encoder-side.
	In addition, \textbf{\textit{ETC Pipeline} BBF} (base/big) outperformed the comparison systems +Localness and +Context-Aware SANs.
	This indicates that additional context (backbone information) is also an effective option for improving the NMT baselines.
	
	4) \textbf{\textit{ETC Pipeline} BBF} (base) is comparable to the +global-deep context, the best comparison system, while \textbf{BBF} (big) slightly outperformed +global-deep context by $0.16$ BLEU score.
	In particular, the parameters of \textbf{\textit{ETC Pipeline} BBF} (base/big) model, which just increased $12.1/7.9$M over the Transformer (base/big), were only 70\% of the +global-deep context model.   
	This shows that the \textbf{\textit{ETC Pipeline} BBF} model is more efficient than the +global-deep context model, though the training speed of the proposed models slightly decreased ($8\%$), compared to the corresponding baselines.
	
	5) The proposed \textbf{\textit{ETC Pipeline} BBF} (base) slightly outperformed Transformer (big), this  contains much more parameters than \textbf{\textit{ETC Pipeline} BBF} (base). This indicates that our improvement is not likely to be due to the increased number of parameters. 
	
	For the EN-FR translation task, the proposed models gave similar improvements over the baseline systems and other top models (though Transformer (big) had a much greater improvement over Transformer (base)). These results show that our method is robust in improving the translation of other language pairs.
	
	\textcolor{black}{In the text compression evaluation, we found ITC did not perform as well as ETC and was even inferior to some other text summarization models; however, since the ITC module will be optimized in the training of downstream tasks, ts performance on only text compression is not necessarily indicative of its performance on the downstream task, so we also conducted experiments on \textbf{\textit{ITC Joint} BBF}.}
	Comparing the results of \textbf{\textit{ITC Joint} BBF}, \textbf{\textit{ETC Pipeline} BBF}, and the baselines, \textit{ITC Joint} brings improvements to the baseline model, which again illustrates the contribution of text compression to machine translation. In addition, \textit{ITC Joint} obtains greater improvements than \textit{ETC Pipeline}, indicating that the joint training can take advantage of the similarities in tasks and reduce error propagation. Besides, the number of parameters in the \textit{ITC Joint} model is not directly comparable to that of the \textit{ETC Pipeline} model as the parameters of the ETC model are not included in the total of the machine translation model. Additionally, the speed of translation did not slow down and even increased, which is due to the fact that ITC eliminates the redundant encoding process and adopts a non-autoregressive decoder instead of an autoregressive decoder.

	\subsection{Machine Reading Comprehension}
	
	\textcolor{black}{In addition to the evaluation on the \textit{encoder-decoder} downstream NMT model, we also conducted experiments with the \textit{encoder-classifier} MRC models.}
	Table \ref{tab:squad2.0} shows the results for reading comprehension on the SQuAD 2.0 dev and test sets\footnote{To focus on the evaluation of ETC and keep simplicity, we only compared with single models instead of ensemble ones.}. ETC solidly improves over the strong BERT baseline in both EM and F1. With the more powerful ALBERT, ETC still obtains improvements and also outperforms all the published works and achieves comparable performance with a few unpublished models from the leaderboard\footnote{The test set needs to be evaluated online, and the results on the test set were not available at the time of submission.}.  In particular, our model is slightly better than the state-of-the-art ALBERT+Retro-Reader\cite{zhang2020retrospective}. This means that text compression is beneficial for the reading comprehension task. \textcolor{black}{In addition, comparing \textit{ETC Pipeline} and \textit{ITC Joint}, the ITC approach definitely benefits from joint training and can generally obtain better performance.}
	
	\begin{table}[t]\small
		\centering
		\caption{\label{tab:squad2.0} \textcolor{black}{Exact Match (EM) and F1 scores on the SQuAD 2.0 dev and test sets for single models.}
		}
		\begin{tabular}{l c c c c}
			\toprule
			\multirow{2}{*}{\textbf{Model} }& \multicolumn{2}{c}{\textbf{Dev}} & \multicolumn{2}{c}{\textbf{Test}}\\
			\cmidrule(lr){2-3} \cmidrule(lr){4-5} &   \textbf{EM} & \textbf{F1}	&   \textbf{EM} & \textbf{F1}\\
			\midrule
			\multicolumn{5}{c}{\emph{Regular Track}} \\ 
			Joint SAN & & & & \\
			U-Net\cite{sun2018u}  & 70.3 & 74.0 & 69.2 & 72.6 \\
			RMR+ELMo+Verifier\cite{hu2019read+} & 72.3 & 74.8 & 71.7 & 74.2\\
			\midrule
			\multicolumn{5}{c}{\emph{Leaderboard (Feb. 23st, 2020)}} \\
			Human & N/A & N/A & 86.83 &	89.45 \\
			BERT &  78.70 & 81.90 & 80.00 & 83.06 \\
			XLNet\cite{yang2019xlnet} & 87.90 & 90.60 & 87.92 & 90.68 \\
			ALBERT\cite{lan2019albert} & 87.40 & 90.20 & 88.10 & 90.90 \\
			ALBERT+Verifier & N/A & N/A & 88.35 &	91.01 \\
			ALBERT+Retro-Reader\cite{zhang2020retrospective} & 87.80 & 90.90 & 88.10 & 91.41 \\
			\midrule
			\multicolumn{5}{c}{\emph{Our implementation}} \\
			BERT Baseline &  78.57 & 81.84  &  $-$ &  $-$  \\
			\quad\textbf{+\textit{ETC Pipeline}} & 79.33  & 82.59  & $-$ & $-$  \\
			\quad\textbf{+\textit{ITC Joint}} & 79.62 & 82.94 & $-$ & $-$\\
			\hdashline
			BERT + Verifier & 79.52  & 82.61  &  $-$ &  $-$  \\
			\quad\textbf{+\textit{ETC Pipeline}} & 79.94 & 83.23  & $-$ & $-$  \\
			\quad\textbf{+\textit{ITC Joint}} & 80.15 & 83.54 & $-$ & $-$\\
			\hdashline
			ALBERT Baseline & 87.00  & 90.15   & $-$  &  $-$  \\
			\quad\textbf{+\textit{ETC Pipeline}} & 87.50 & 90.50 & $-$ & $-$ \\
			\quad\textbf{+\textit{ITC Joint}} & 87.75 & 90.85 & $-$ & $-$ \\
			\hdashline
			ALBERT + Verifier & 87.42 & 90.45  & $-$  &  $-$  \\
			\quad\textbf{+\textit{ETC Pipeline}} & 87.98 & 90.97 & $-$ & $-$ \\
			\quad\textbf{+\textit{ITC Joint}} & 87.95 & 91.00 & $-$ & $-$ \\
			\bottomrule
		\end{tabular}
	\end{table}
	
	Table~\ref{tab:race} shows the results on the RACE task. The re-implemented BERT and ALBERT baselines performed comparably to the \emph{Leaderboard}, indicating that they are strong baselines. Our model boosted the BERT baseline substantially, with an increase of +0.8\% accuracy, and it boosted the ALBERT baseline with +0.6\% accuracy on the test set. This improvement further verifies that the proposed method is a robust method for improving performance on the MRC task.
	In addition, although there is still a gap between our model and the state-of-the-art ALBERT+DUMA\cite{zhu2020dual}, which improves the performance with a more complicated neural network structure, on the current leaderboard, we obtain improvement from the representation learning, which is orthogonal to their improvement, so our results could be further improved through their complicated network structure. \textcolor{black}{Similar to its results on the SQuAD task, \textit{ITC Joint} outperforms ETC Pipeline on the RACE task. These results indicate that although the compressed text obtained by ETC is better than the ITC for text compression on its own with automated metrics such as F1, its results are not as good as those of ITC on the downstream task. This may shows that the backbone information for different tasks is not static. Finetuning text compression and downstream tasks together can lead to more task-relevant backbone information, which can better improve the performance of downstream tasks.}
	
	\begin{table}[htbp]\small
    	\caption{\textcolor{black}{Accuracy on the RACE test set for single models.}}
    	\label{tab:race}
    	\centering
    		\begin{tabular}{l c c c}
    			\toprule
    			Model &  \textbf{Middle} & \textbf{High} & \textbf{All}	\\
    			\hline
    			\multicolumn{4}{c}{\emph{Human Performance}} \\
    			Turkers  & 85.1 & 69.4 & 73.3\\
    			Ceiling  & 95.4 & 94.2& 94.5 \\
    			\midrule
    			\multicolumn{4}{c}{\emph{Leaderboard (Feb. 23st, 2020)}} \\
    			BERT \cite{ran2019option} & 75.6 & 64.7 & 67.9  \\
    			XLNet\cite{yang2019xlnet} & 85.5 & 80.2 & 81.8  \\
    			ALBERT\cite{lan2019albert} & 89.0 & 85.5 & 86.5 \\ 
    			ALBERT+DUMA\cite{zhu2020dual} & 90.9 & 86.7 & 88.0 \\
    			\midrule
    			\multicolumn{4}{c}{\emph{Our implementation}} \\
    			BERT Baseline & 76.6 & 70.1  & 72.0 \\
    			\quad\textbf{+\textit{ETC Pipeline}}   & 77.8 & 70.7  & 72.8 \\
    			\quad\textbf{+\textit{ITC Joint}} & 78.3 & 70.9 & 73.4 \\
    			\hdashline
    			ALBERT Baseline & 88.7 & 85.6 & 86.5 \\
    			\quad\textbf{+\textit{ETC Pipeline}}  & 89.3 & 86.2 & 87.1 \\
    			\quad\textbf{+\textit{ITC Joint}} & 89.6 & 86.3 & 87.3 \\
    			\bottomrule
    		\end{tabular}
    \end{table}

	\section{Ablation Study}\label{sec:ablation}
	
	\subsection{Effect of Different Levels of Text Compression Quality}
	
	To show effect of different levels of text compression quality on the improvement of downstream tasks, we conducted experiments on the WMT14 EN-DE translation task using Transformer (base) \textbf{+\textit{ETC Pipeline} BBF} and \textbf{+\textit{ITC Joint} BBF}. 
	We considered six different levels of compression quality resulting from the following settings: AllText, F8W, RandSample (random sampling), Supervised, Unsupervised, and Semi-supervised. The performance comparison is shown in Table \ref{tab:KC_performance}.
	
	\begin{table}[h]
		\centering
		\caption{\textcolor{black}{The effect of levels of text compression quality on translation baseline. The results are reported on the newstest2014 test set.} \label{tab:KC_performance}}
		\begin{tabular}{lcc}
			\toprule
			Model & \textit{ETC Pipeline} & \textit{ITC Joint} \\ 
			\midrule
			Baseline            &   27.24   & 27.24 \\ 
			\quad+\textbf{AllText}  & 27.24 & $-$\\
			\quad+\textbf{F8W}  & 27.40 & $-$\\
			\quad+\textbf{RandSample} & 26.53 & $-$\\
			\quad+\textbf{Supervised}   & 27.80  & 27.64 \\
			\quad+\textbf{Unsupervised} &   27.97 & 28.10 \\ 
			\quad+\textbf{Semi-supervised}   &   28.35  & 28.57 \\
			\bottomrule
		\end{tabular}
	\end{table}  
	
	We made the following observations: 
	
	1) Simply introducing AllText and F8W achieved few improvements, and RandSample was outperformed by the baseline. In comparison, the  +Supervised, +Unsupervised, and +Semi-supervised models substantially improved the performance over the baseline Transformer (base).
	This means that our text compression approach provides richer and useful source information for machine translation tasks.
	
	2) +Unsupervised models can lead to better improvements compared to +Supervised models, although supervised models outperform unsupervised models in the text compression benchmark.  This may be due to the fact that the annotated text compression training data is in a domain different from the WMT EN-DE training data. +Semi-supervised models with annotated data finetuning outperformed both +Unsupervised and +Supervised models.
	
	\textcolor{black}{3) Comparing \textit{ETC Pipeline} and \textit{ITC Joint}, \textit{ETC Pipeline} is more effective under a supervised setting, while \textit{ITC Joint} performs better under other settings. This may be because in the case of limited (annotated) training data, \textit{ETC Pipeline} can be optimized better by using an independent compression model, but in {unsupervised} and {semi-supervised} settings, \textit{ETC Pipeline} loses this advantage as there is more data available.}
	
	\subsection{Effect of Encoder Parameters}
	
	In \textit{ETC Pipeline}, representations of the original sentence and its compressed sequence were learned by a shared encoder.
	To explore the effect of the encoder parameters, we also designed a \textbf{BBF} model with two independent encoders to learn respective representations of the original sentence and its compressed version. 
	Table~\ref{tab:encoder_papameter} shows results on the newstest2014 test set for the WMT14 EN-DE translation task.
	
	\begin{table}[h]
		\centering
		\caption{The effect of encoder parameters.}\label{tab:encoder_papameter}
		\begin{tabular}{lcc}
			\toprule
			Model & BLEU & \#Params\\
			\midrule
			Transformer (base) & 27.24  & 66.4M \\ \hdashline
			\textbf{+\textit{ETC Pipeline} BBF w/ Shared encoder} & 28.35 & 78.6M\\  
			\textbf{+\textit{ETC Pipeline} BBF w/ Independent encoders}  &  28.50 & 91.6M\\
			\bottomrule
		\end{tabular}
	\end{table} 
	
	The \textbf{BBF} (w/ independent params) slightly outperformed the proposed shared encoder model by a BLEU score of 0.15, but its parameters increased by approximately 30\%.
	In contrast, the parameters in our model are comparable to those of Transformer (base).
	To account for the number of parameters, we used a shared encoder to learn source representations, which makes it easy to verify the effectiveness of the additional backbone information.
	
	\subsection{Different Compression Ratio Evaluation}
	
	In order to verify the impact of different compression ratios on the downstream task, we conducted experiments on the WMT14 EN-DE translation task with \textit{ETC Pipeline} BBF and \textit{ITC Joint} BBF using Transformer (base) models under a semi-supervised text compression training setting. We gradually increased the compression ratio $\gamma$ from 0 to 1.0. When the compression ratio $\gamma = 0$, no compression sequence is generated, which is the same as the vanilla Transformer. Setting the compression ratio $\gamma = 1.0$ is equivalent to re-paraphrasing the source sentence (maintaining the same length). 
	
	\begin{figure}[thb!]
		\setlength{\abovecaptionskip}{0pt}
		\begin{center}
			\pgfplotsset{height=5.0cm,width=4.0cm,compat=1.14,every axis/.append style={thick}}
			\begin{tikzpicture}
			\begin{axis}
			[width=8.0cm, enlargelimits=0.13, tick align=outside, legend style={cells={anchor=west},legend pos=south east,every axis legend/.append style={
					at={(1,0)}}}, xticklabels={ $0$, $0.1$,$0.2$, $0.3$, $0.4$, $0.5$, $0.6$, $0.7$, $0.8$, $0.9$, $1.0$}, 
			xtick={0,1,2,3,4,5,6,7,8,9,10}, 
			ylabel={BLEU score},xlabel={Compression Ratio},font=\small]
			
			\addplot+ [sharp plot, mark=*,mark size=1.2pt,mark options={solid,mark color=orange}, color=orange] coordinates
			{(0,27.24)(1,27.30)(2,27.41)(3,27.65)(4,27.98)(5,28.31)(6,28.35)(7,28.24)(8,28.01)(9,27.77)(10,27.62)};
			\addlegendentry{\tiny \textbf{ETC Pipeline}}
			\addplot+ [sharp plot, mark=square,mark size=1.2pt,mark options={solid,mark color=red}, color=red] coordinates
			{(0,27.24)(1,27.52)(2,27.94)(3,28.36)(4,28.57)(5,28.48)(6,28.40)(7,28.35)(8,28.10)(9,27.80)(10,27.63)};
			\addlegendentry{\tiny \textbf{ITC Joint}}
			\end{axis}
			\end{tikzpicture}
			\caption{\label{fig:gamma} \textcolor{black}{Performances on EN-DE newstest2014 with different text compression ratios.}}
		\end{center}
	\end{figure}
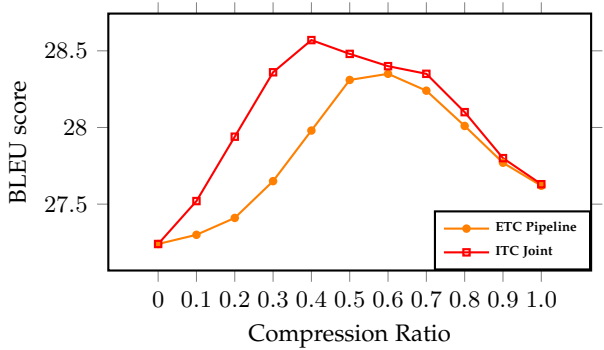

	The experimental results are shown in Fig. \ref{fig:gamma}. As can be seen from the results, text compression can bring performance improvement. When the compression ratio $\gamma=1.0$ and the sentence length is not shortened, re-paraphrasing can still bring a slight improvement in translation quality. \textcolor{black}{Additionally, the experimental results show that the best compression ratio of \textit{ITC Joint} is lower than that of \textit{ETC Pipeline}. This may be because ITC and the downstream tasks are optimized together, and information that is not relevant to the specific downstream task is more often excluded from the compression, which leads to a lower compression ratio in \textit{ITC Joint} achieving the best performance compared with \textit{ETC Pipeline}.}
	
	\subsection{\textcolor{black}{ETC Pipeline, ETC Joint, ITC Joint, and DAE pre-training}}
	
	\textcolor{black}{\textit{ETC Pipeline}, \textit{ETC Joint}, and  \textit{ITC Joint} are the three compression methods we studied. In this section, we compared their performance, training time, and inference speed to visually demonstrate their differences. The first stage of training in \textit{ITC Joint} is a bit similar to DAE pre-training of a model's encoder, so we take this as one of the baselines for comparison. We only compare the actual downstream model training and inference speed. The time in pre-training and compression generation are not considered. The baseline model is still Transformer (base), and the evaluation task is WMT14 EN-DE translation. All models are trained for 20w steps to make the training time comparable, and the results are shown in Table \ref{tab:etc_itc}.}
	
	\begin{table}[h]
		\centering
		\caption{\textcolor{black}{Comparison of \textit{ETC Pipeline}, \text{ETC Joint}, \textit{ITC Joint}, and DAE pre-training approaches.} \label{tab:etc_itc}}
		\begin{tabular}{lccc}
			\toprule
			Systems & BLEU & \#Training & \#Speed \\
			\midrule
			Transformer (base) &  27.24 & 16h & 131k \\
			\quad+DAE & 27.65 & 16h & 131k \\
			\hdashline
			\quad\textbf{+\textit{ETC Pipeline} BBF} & 28.35 & 18h &  119k \\ 
			\quad\textbf{+\textit{ETC Joint} BBF} & 28.04 & 46h & 65k \\ 
			\quad\textbf{+\textit{ITC Joint} BBF} & 28.57 & 20h & 120k \\ 
			\bottomrule
		\end{tabular}
	\end{table} 
	
	\textcolor{black}{The comparison results show that the training time of \textit{ETC Joint} is much longer than those of \textit{ETC Pipeline} and \textit{ITC Joint} because of the autoregressive decoding for each batch in training. Secondly, although the \textit{ETC Joint} outperforms the baseline, the improvement is less than that of the \textit{ETC Pipeline}. The reason may be that only the self attention part in the ETC Model is jointly updated in the downstream task training, which is no better than re-encoding using the trainable encoder in the downstream model. Also, the DAE pre-training method can improve the baseline's performance to a certain extent, but the improvement is less than that of our text compression, which verifies that does not only rely on pre-training, and backbone information is indeed useful.}
	
	\section{Further Exploration}

	\subsection{Transfer Learning Effects of ETC}
	
	In our experiments on machine translation, we found that the semi-supervised setting was best when enhancing the baseline NMT model with ETC. We hypothesize that this results from the inconsistent domains of the training data in machine translation and text compression. To verify this hypothesis, we carried out a domain adaptation experiment on the IWSLT 2015 \cite{cettolo2015iwslt} EN-DE task.
	
	The training data of IWSLT 2015 EN-DE is taken as the in-domain (TED domain) training set for the experiment, and TED tst2010 is used as the development set. TED tst2011, tst2012, tst2013 are concatenated to make the in-domain test set. The WMT 14 EN-DE newstest2014 is used as the out-of-domain test set. We use the Transformer (base) \textbf{\textit{ETC Pipeline} BBF} model of the previous experiment for the NEWS domain. For the TED domain, we finetune the NEWS domain ETC model using the source corpus of IWSLT15 EN-DE with an unsupervised text compression target.
	
	\begin{table}[h]
		\centering
		\caption{The transfer learning results of ETC-aided models. \label{tab:transfer_learning}}
		\begin{tabular}{lccc}
			\toprule
			\multirow{2}{*}{Systems} & \multirow{2}{*}{ETC domain} & IWSLT & WMT \\
			 & & (TED) & (NEWS) \\
			\midrule
			Transformer (base) & - & 30.76  & 5.24 \\ \hdashline
			\multirow{2}{*}{\quad\textbf{+\textit{ETC Pipeline} BBF}} & NEWS & 30.54 &  7.88 \\  
			& TED & 31.28 & 4.91 \\  
			\bottomrule
		\end{tabular}
	\end{table} 
	
	\begin{table*}[h]
		\centering
		\caption{Accuracy results on the MNLI dev dataset, including match (m) and mismatch (mm) parts, and results on the HANS dataset.}\label{tab:mnli} 
		\begin{tabular}{lccccccc}
			\toprule
			& \textbf{MNLI} & \multicolumn{3}{c}{\textbf{Heuristic Entailed}} & \multicolumn{3}{c}{\textbf{Heuristic Non-entailed}}\\ 
			& m/mm(acc) & lexical overlap &  subsequence & constituent & lexical overlap &  subsequence & constituent \\
			\midrule
			BERT$_\text{base}$ & 84.61/83.43 & 93.04 & 97.76 & 96.82 & 59.90 & 3.96 & 11.80 \\
			\hdashline
			BERT$_\text{large}$ & 86.24/85.96 & 97.02 & 99.42 & 99.62 & 70.40 & 29.08 & 33.48  \\
			\textbf{+\textit{ETC Pipeline} BEF} & 86.69/86.35 & 98.10 & 99.22 & 99.48  &  75.46  & 29.22 & 34.12 \\
			\hdashline
			ALBERT$_\text{xxlarge}$ & 90.62/90.59 & 100 & 99.92  & 99.98 &  88.48  & 29.86  & 10.34 \\
			\textbf{+\textit{ETC Pipeline} BEF} & 90.85/90.74 & 99.90 & 99.96 & 99.72 & 94.24  & 35.42 & 14.80 \\
			\bottomrule
		\end{tabular}
	\end{table*}
	
	Due to the small scale of the IWSLT15 training set, we choose Transformer (base) as the baseline. The experimental results are shown in Table \ref{tab:transfer_learning}. \textbf{+\textit{ETC Pipeline} BBF} brings 0.52 and 2.64 BLEU points improvements over its baseline model when the domain of ETC model is the same as the test set. 
    This shows the proposed ETC method has the function of transfer learning to a certain extent when the domain of ETC model is the same as the test set. We suspect that this is because the ETC model can extract a sentence's useful backbone information. Although the training and test sets have different domains and different styles, the backbone structure of a sentence is a consistent property that can benefit the model in different domains. \textbf{+\textit{ETC Pipeline} BBF} enhances the robustness of the model by normalizing the changes caused by different domains.
    
    In addition, when the domain of the ETC model is inconsistent with the test set, the results on the test set show some decline compared to the baseline. The reason for this phenomenon may be due to the inconsistency in domain, which the compression generated to worsen, and the error propagates to downstream models. In this case, finetuning the ETC model using the monolingual corpus of the target domain is an effective and necessary operation.
	
    \subsection{Language Representation Improvement Contribution}
    
	To evaluate the contributions of the language representation improvement in the text compression-aided models, we performed an ablation study on an NLI task. NLI involves reading a pair of sentences $\langle \textit{premise}, \textit{hypothesis}\rangle$ and judging the relationship between their meanings, such as entailment, neutral, or contradiction. We evaluated our ETC-aided model on and alongside a strong ALBERT baseline and train both the ETC-aided model and the baseline on the  MNLI dataset using adversarial evaluation of natural language inference with the Heuristic Analysis for the NLI Systems (HANS) dataset proposed by McCoy et al. \cite{mccoy-etal-2019-right}. 
	
	McCoy et al. \cite{mccoy-etal-2019-right}, aiming to give the model some hints during inference, suggest using three heuristics for good performance for models in NLI:
	\begin{itemize}
		\item \textbf{lexical overlap}: assume that the premise entails all hypotheses constructed from words in the premise.
		\item \textbf{subsequence}: assume that the hypothesis is the substring of premise. 
		\item \textbf{constituent}: assume that the premise entails all complete subtrees in its parse tree.
	\end{itemize}
	
	The instances labeled ``entailed" and ``non-entailed" in HANS that meet with the three heuristics, obey a uniform distribution. Table \ref{tab:mnli} shows results on the MNLI dev dataset and HANS dataset. From the results, we can see that:
	
	 1) The performance of models on ``entailed" is better than on ``non-entailed." This suggests that the model is making excessive use of heuristic information. According to McCoy et al.'s \cite{mccoy-etal-2019-right} explanation, MNLI training data might not provide not enough supervision to learn the desired level of NLI.
	 
	 2) From BERT$_\text{base}$ to BERT$_\text{large}$ to ALBERT$_\text{xxlarge}$, the model parameters gradually increased, and the results on the MNLI dev set improved. The results of entailed and non-entailed data were generally better, but the result of BERT on non-entailed ``constituent" data was stronger than that of ALBERT. This may be due to the loss of syntactic information due to the sharing of ALBERT layer parameters.
	 
	 3) The ETC-aided model we proposed can enhance performance on ``non-entailed" data, indicating that the quality of language representation has been improved.

	 \subsection{Probing Evaluation for the Language Representation}
	 
	 The assessment of the MNLI model on the HANS dataset shows that the improvement of results does come from a better quality language representation rather than tricks. In order to understand what parts of the language representations improved, we further evaluate the MNLI model using ten widely-used language probing tasks \cite{conneau-etal-2018-cram}.
	 
	 Specifically, we use the BERT$_{base}$ and ALBERT$_{base}$ models trained for the MNLI task to generate a sentence representation (embedding) of the inputs to evaluate the quality of the linguistic properties are encoded in them. The evaluation results are shown in Table \ref{table:probing_results}. Comparing the two baselines, BERT and ALBERT, shows that ALBERT uses only about 1/10 the parameters of BERT and achieves a comparable results. \textbf{+\textit{ETC Pipeline} BEF} achieves statistically significant improvement on the WC, TopConst, and  Tense tasks compared to the two baselines, which demonstrates that the ETC-aided model focuses on important words, constituents, and  verbs more in its sentence representations, which is consistent with our hypothesis.
	 
	\begin{table*}[t]
		\caption{Ten probing task accuracies.  "++/+" after the accuracy score indicate that the score is statistically significant at level $p < 0.01/0.05$. \textcolor{black}{The tasks are abbreviated: SentLen: sentence length, WC: important word, TreeDepth: syntactic tree depth, TopConst: top constituents, BShift: word order, Tense: verb tense, SubjNum: subject number, ObjNum: object number, SOMO: semantic replacement, CoordInv: coordinated clausal conjoints order.}}\label{table:probing_results}
		\resizebox{1\linewidth}{!}{
			\begin{tabular}{@{}l@{\,}ccccccccccc}
				\toprule
				\bf Task & \bf SentLen & \bf WC & \bf TreeDepth & \bf TopConst & \bf BShift & \bf Tense & \bf SubjNum & \bf ObjNum &  \bf SOMO & \bf CoordInv & \bf \#Params \\
				\midrule
				BERT$_\text{base}$ & 65.84 &  50.35 & 29.97 & 53.31 & 72.22 & 86.35 & 77.63 & 76.23 & 61.18 & 61.14 & 109M \\
				\quad\textbf{+\textit{ETC Pipeline} BEF} & 65.78 & \quad59.24++ & 30.01 & \quad55.42++ & 71.79 &\quad86.83+$\ $ & 77.79 & 76.58 & 60.78 & 60.06 & 110M \\
				\hdashline
				ALBERT$_\text{base}$ & 65.46 & 50.02 & 31.11 & 52.99 & 71.82 & 86.28 & 77.89 & 76.51 & 60.36 & 60.34 & 12M \\
				\quad\textbf{+\textit{ETC Pipeline} BEF} & 65.45 & \quad58.78++ & 31.02 & \quad55.07++ & 71.15 &\quad86.68+$\ $ & 77.95 & 76.81 & 59.98 & 60.41 & 13M \\
				\bottomrule
			\end{tabular}
		}
		\vspace{-0.4cm}
	\end{table*}

	\subsection{ETC Enhancement on Different MAX\_SEQ\_LEN}
	
	In Transformer encoders, the $O(N^2)$ self-attention operation is a process where every output element is connected to every input element, and the weightings between them are dynamically calculated based upon the circumstances. Transformer encoders are very flexible compared to models with fixed connectivity patterns because of this nature, but they consume large amounts of memory and inference time when applied to data types with many elements, such as the MRC task.
	
	Currently, there is an upper limit on the memory of a single NVIDIA GPU, so the mainstream approach to reducing the memory requirement for a long sequence is to truncate long sentences or adopt a sliding window with a limited window size for them. The first method is typically used for choice-style MRC tasks (RACE), and the second is used for span-style MRC tasks (SQuAD); however, both methods have an impact on the performance of our model. 
	
	To explore the effect of ETC enhancements at different maximum sequence lengths (MAX\_SEQ\_LEN), we report the results of the ALBERT baseline and our proposed ETC-aided model on the SQuAD 2.0 development set and the RACE test set, as shown in Table \ref{tab:squad2.0_msl}. The experimental results show a downward trend in model performance with a decrease of the maximum sequence length. Under different MAX\_SEQ\_LEN, the performance improvements on SQuAD are 0.52, 0.46 and 0.50, and 0.6, 0.8 and 0.9 on RACE, respectively. Adopted by SQuAD to predict the answer position, the sliding window mechanism alleviates the problem caused by the maximum length, and thus, the relationship between ETC and the magnitude of performance variation could not be clearly seen. In the RACE task, the model truncates length over MAX\_SEQ\_LEN. With this reduction of the maximum length, the performance degradation is reduced, showing that ETC improves the quality of truncated text representations.
	
	\begin{table}[thtbp]\small
		\centering
		\setlength{\tabcolsep}{3pt}
		\caption{\label{tab:squad2.0_msl} Results on the SQuAD2.0 dev and RACE test sets with different MAX\_SEQ\_LEN (MSL). M./H. represent middle and high splits of the RACE test set.}
		\begin{tabular}{l c c c c c }
			\toprule
			\multirow{2}{*}{\textbf{Model}} & \multirow{2}{*}{\textbf{MSL}} & \multicolumn{2}{c}{SQuAD} & \multicolumn{2}{c}{RACE}	\\
			\cmidrule(lr){3-4} \cmidrule(lr){5-6} & & \textbf{EM} & \textbf{F1} & \textbf{M./H.} & \textbf{All} \\
			\midrule
			ALBERT Baseline &  512  & 87.42  &  90.45  & 88.7/85.6 & 86.5 \\
			\quad\textbf{+\textit{ETC Pipeline} BEF} &  512  & 87.98 & 90.97 & 89.3/86.2 & 87.1 \\
			\hdashline
			ALBERT Baseline &  384  & 86.87  &  89.98 & 87.9/84.2 & 85.2 \\
			\quad\textbf{+\textit{ETC Pipeline} BEF} &  384  & 87.36 & 90.44 & 88.7/84.9 & 86.0 \\
			\hdashline
			ALBERT Baseline &  256  & 86.33  &  89.51 & 87.0/82.6 & 83.8 \\
			\quad\textbf{+\textit{ETC Pipeline} BEF} &  256  & 86.85 & 90.01 & 87.7/83.6 & 84.7  \\
			\bottomrule
		\end{tabular}
	\end{table}

	\section{Related Work}\label{sec:related}
	
	Exploiting sentence segmentation, sentence simplification, and text compression have all been used to emphasize the source sentence information for machine translation. \cite{xiao-etal-2014-hybrid} presented an approach to integrating the sentence skeleton information into a phrase-based statistic machine translation system. \cite{xiao2016syntactic} proposed an approach to modeling syntactically-motivated skeletal structure of a source sentence for statistic machine translation.
	\cite{mellebeek2006syntactic} described an early approach to skeleton-based translation, which decomposes input sentences into syntactically meaningful chunks. The central part of the sentence is identified and remains unaltered while other parts of the sentence are simplified. This process produces a set of partial, potentially overlapping translations that are recombined to form the final translation. 
	\cite{sudoh-etal-2010-divide} described a ``divide and translate" approach to dealing with complex input sentences. They parse the input sentences, replace subclauses with placeholders, and later substitute them with separately translated clauses. Their method requires training translation models on clause-level aligned parallel data with placeholders in order for the translation model to deal with the placeholders correctly.
	\cite{pouget-abadie-etal-2014-overcoming} experimented with automatically segmenting the source sentence to overcome problems with overly long sentences. 
	\cite{hasler2017source} showed that the spaces of original and simplified translations could be effectively combined using translation lattices and compared two decoding approaches to process both inputs at different levels of integration. 
	
	Different from these works, our proposed text compression model does not rely on any known linguistics-motivated (such as syntax) skeleton simplification but directly trains a computation-motivated text compression model to learn to compress sentences and re-paraphrase them directly in a seq2seq model. Our text compression model can surprisingly generate more grammatically correct and refined sentences, and the words in the compressed sentence do not have to be the same as those in the original sentence. At the same time, our text compression model can give source backbone representation exempt from the unstable performance of a syntactic parser, which is especially important as stable performance is essential for syntactic skeleton simplification. Our text compression model can perform unsupervised training on large-scale datasets and then use the supervised data for finetuning, which produces more impressive results.
	
	\section{Conclusion}\label{sec:conclusion} 
	
	This paper presents explicit and implicit text compression methods for improving representations from Transformer encoders. We aim to use the backbone knowledge of text to generate more accurate characteristic statistical distributions in language representation learning. To demonstrate that the proposed text compression enhancement serves a general purpose in NLP tasks, we evaluate the impact of the proposed approach in two major NLP tasks, machine translation and machine reading comprehension, and in an extra natural language inference task for natural language understanding evaluation. The experimental results show that our proposed model can yield significant improvements over strong baselines on flagship datasets and benchmark leaderboards for these challenging tasks.

	\ifCLASSOPTIONcompsoc
	% The Computer Society usually uses the plural form
	\section*{Acknowledgments}
	\else
	% regular IEEE prefers the singular form
	\section*{Acknowledgment}
	\fi
	We thank Kevin Parnow, from the Department of Computer Science and Engineering, Shanghai Jiao Tong University (parnow@sjtu.edu.cn) for editing a draft of this manuscript.

	% Can use something like this to put references on a page
	% by themselves when using endfloat and the captionsoff option.
	\ifCLASSOPTIONcaptionsoff
	\newpage
	\fi
	
	\bibliographystyle{IEEEtran}
	\bibliography{references}

\end{document}